\documentclass{article} 

\usepackage{spconf_one}

\usepackage{amsmath}
\usepackage{graphicx}
\usepackage{tabularx}
\usepackage{epstopdf}
\usepackage{my_symbols}
\usepackage{subfig}
\usepackage[lined]{algorithm2e}
\usepackage{enumitem}

\newcommand{\green}{\textcolor{green}}		

\newcommand{\alg}[1]{\texttt{#1}}		

\makeatletter
\newcommand{\removelatexerror}{\let\@latex@error\@gobble}
\makeatother

\usepackage{tabularx}

\usepackage{verbatim}

\begin{document}
%

\title{Scalable Multi-Output Label Prediction:\\From Classifier Chains to Classifier Trellises}



\name{J. Read$^1$, L. Martino$^2$, P. M. Olmos$^3$, David Luengo$^4$}
\address{$^1$ Dep.\ of Computer Science, Aalto University and HIIT, Helsinki, Finland (\url{jesse.read@aalto.fi}). \\
$^2$ Dep. of Mathematics and Statistics, University of Helsinki, Helsinki (Finland).\\
$^3$ Dep.\ of Signal Theory and Communications, Universidad Carlos III de Madrid (Spain) .\\
$^4$ Dep. of Circuits and Systems Engineering, Universidad Polit\'ecnica de Madrid,  (Spain).
}

\maketitle

\begin{abstract}
Multi-output inference tasks, such as multi-label classification, have become increasingly important in recent years.
A popular method for multi-label classification is \emph{classifier chains}, in which the predictions of individual classifiers are cascaded along a chain, thus taking into account inter-label dependencies and improving the overall performance.
Several varieties of classifier chain methods have been introduced, and many of them perform very competitively across a wide range of benchmark datasets.
However, scalability limitations become apparent on larger datasets when modeling a fully-cascaded chain.
In particular, the methods' strategies for discovering and modeling a good chain structure constitutes a mayor computational bottleneck.
%
%
In this paper, we present the \key{classifier trellis} (CT) method for scalable multi-label classification. We compare CT with several recently proposed classifier chain methods to show that it occupies an important niche: it is highly competitive on standard multi-label problems, yet it can also scale up to thousands or even tens of thousands of labels.
\end{abstract}




\section{Introduction}
\label{sec:Introduction}

\begin{figure}[hb]
    \centering 
\begin{center}
		 \includegraphics[scale=0.31]{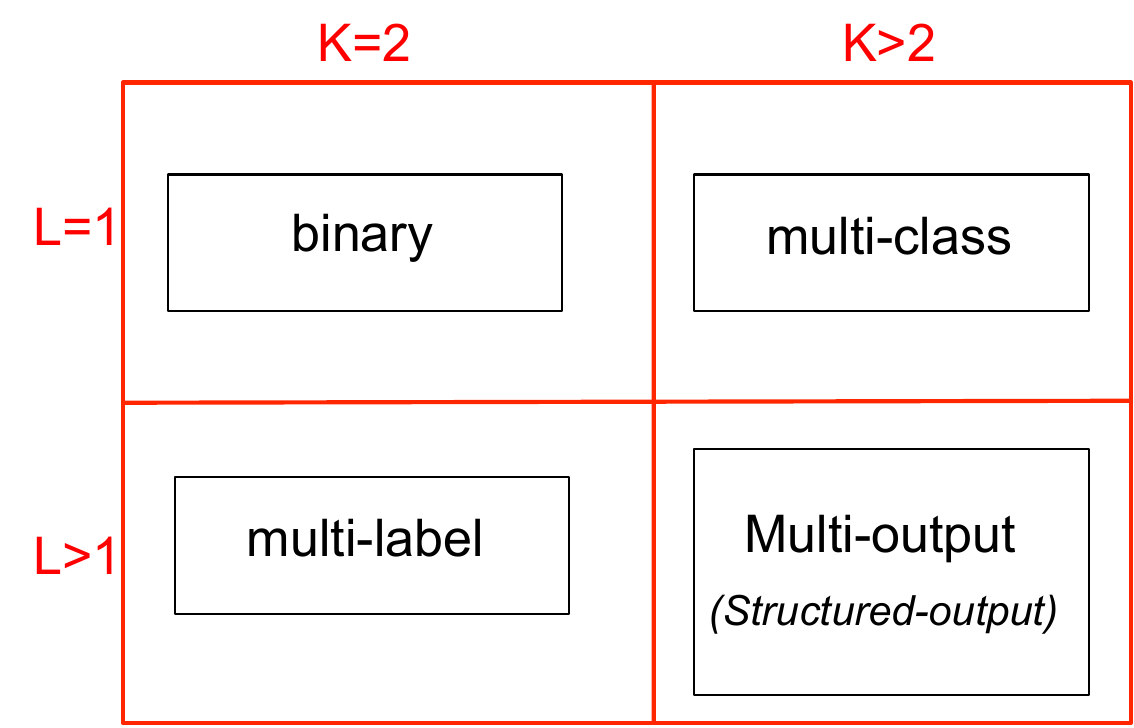}
	\end{center}
	\caption{\label{fig:intro} Different classification paradigms: $L$ is the number of \emph{labels} and $K$ is the number of \emph{values} that each label variable can take.
	} 
\end{figure}

Multi-output classification (MOC) (also known variously as multi-target, multi-objective, and multidimensional classification) is the supervised learning problem where an instance is associated with a set of qualitative discrete variables (a.k.a. \emph{labels}), rather than with a single variable
\footnote{We henceforth try to avoid the use of the term `class'; it generates confusion since it is used variously in the literature to refer to both the target variable, and a value that the variable takes. Rather, we refer to \emph{label} variables, each of which takes a number of \emph{values}.}.
Since these label variables are often strongly correlated, modeling the dependencies between them allows MOC methods to improve their performance at the expense of an increased computational cost. Multi-label classification (MLC) is a special case of MOC where all the labels are binary; it has already attracted a great deal of interest and development in machine learning literature over the last few years. In \cite{Review}, the authors give a recent review of, and many references to, a number of recent and popular methods for MLC. 
\Fig{fig:intro} shows the relationship between different classification paradigms, according to the number of labels ($L=1$ vs. $L>1$) and their type (binary $[K=2]$ or not $[K>2]$).



There are a vast range of active applications of MLC, including tagging images, categorizing documents, and labelling video and other media, and learning the relationship among genes and biological functions. Labels (e.g., tags, categories, genres) are either relevant or not. For example, an image may be labelled \texttt{beach} and \texttt{urban}; a news article may be sectioned under \texttt{europe} and \texttt{economy}. Relevance is usually indicated by $1$, and irrelevance by $0$. The general MOC scheme may add other information such as month, age, or gender. Note that month $\in \{1,\ldots,12\}$ and therefore is not simply irrelevant or not.
This MOC task has received relatively less attention than MLC (although there is some work emerging, e.g., \cite{BCC} and \cite{MCC2}). However, most MLC-transformation methods (e.g., treating each label variable as a separate multi-class problem) are equally applicable to MOC. Indeed, in this paper we deal with a family of methods based on this approach. Note also that, as any integer can be represented in binary form (e.g., $3 \Leftrightarrow [0,0,0,1,1]$), any MOC task can `decode' into a MLC task and vice versa.  

In this paper, we focus on \emph{scalable}  MLC methods, able to effectively deal with large datasets at feasible complexity. Many recent MLC methods, particularly those based on classifier chains, tend to be over engineered, investing evermore computational power to model label dependencies, but presenting poor scalability properties. In the first part of the paper, we  review some state-of-the-art methods from the MLC and MOC literature to show that the more powerful solutions are not well-suited to deal with  large-size label sets. For instance,  \emph{classifier chains} \cite{PCC,ECC2} consider  a full cascade of labels along a chain to model their joint probability distribution and, either they explore all possible label orders in the chain, incurring in exponential complexity with the number of labels, or they compare a small subset of them chosen at random, which is ineffective for large dimensional problems.

The main contribution of the paper is a novel highly-scalable method: the \emph{classifier trellis} (CT). Rather than imposing a long-range and ultimately computationally complex dependency model, as in classifier chains, CT captures the essential dependencies among labels very efficiently. This is achieved by considering a predefined trellis structure for the underlying graphical model, where dependent nodes (labels) are sequentially placed in the structure according to easily-computable probabilistic measures. Experimental results across a wide set of datasets show that CT is able to scale up to large sets (namely, thousands and tens of thousands of labels) while remaining very competitive on standard MLC problems. In fact, in most of our experiments, CT was very close to the running time of the naive baseline method, which neglects any statistical dependency between labels. Also, an ensemble version of CT, where the method is run multiple times with different random seeds and classification is done through majority voting, does not significantly outperform the single-shot CT. This demonstrates that our method is quite robust against initialization.

The paper is organized as follows. First, in \Sec{sec:notation} we formalize the notation and describe the problem's setting. In \Sec{sec:prior} we review some state-of-the-art methods from the MLC and MOC literature, as well as their various strategies for modeling label dependence. This review is augmented with empirical results.
In \Sec{sec:CT} we make use of the studies and theory from earlier sections to present the classifier trellis (CT) method. In \Sec{sec:experiments} we carry out two sets of experiments: firstly we compare CT to some state-of-the-art multi-label methods on an MLC task; and secondly, we show that CT can also provide competitive performance on typical structured output prediction task (namely localization via segmentation). Finally, in \Sec{sec:conclusions} we discuss the results and take conclusions.

Two appendixes have been included to help the readability of the paper and support the presented results.  In \ref{sec:appendixA}, we compare two low-complexity methods to infer the label dependencies from training data. In \ref{sec:inference} we review Monte Carlo methods, which are required in this paper to perform probabilistic approximate inference of the label set associated to a new test input.

\section{Problem Setup and Notation}
\label{sec:notation}
Following a standard machine learning notation, the $n$-th feature vector can be represented as
\begin{equation*}
	\x^{(n)} = [x_1^{(n)},\ldots,x_D^{(n)}]^{\top}
		\in {\boldsymbol \X} = \mathcal{X}_1 \times \cdots \times \mathcal{X}_D \subseteq \mathbb{R}^D,
\end{equation*}
where $D$ is the number of features and $\mathcal{X}_d$ ($d=1,\ldots,D$) indicates the support of each feature.
In the traditional \key{multi-class classification} task, we have a single target variable which can take \emph{one} out of $K$ values, \ie
$$
    y^{(n)} \in  \mathcal{Y}=\{1,\ldots,K\},
$$
and for some test instance $\xtest$ we wish to predict 
\begin{equation}
	\yp = h(\xtest) = \argmax_{y \in  \mathcal{Y}} p(y|\xtest),
\label{eq:MC}
\end{equation}
in such a way that it coincides with the true (unknown) test label with a high probability\footnote{Eq. \eqref{eq:MC} corresponds to the widely used maximum a posteriori (MAP) estimator of $\yt$ given $\xtest$, but other approaches are possible.}.
Furthermore, the conditional distribution $p(y|\x)$ is usually unknown and has to be estimated during the classifier construction stage.
In the standard setting, classification is a supervised task where we have to infer the model $h$ from a set of $N$ labelled examples (training data) $\D = \{(\x^{(n)},y^{(n)})\}_{n=1}^N$, and then apply it to predict the labels for a set of novel unlabelled examples (test data).
This prediction phase is usually straightforward in the single-output case, since only one of $K$ values needs to be selected.

In the \key{multi-output classification} (MOC) task, we have $L$ such output labels,
$$
	\y^{(n)} = [y^{(n)}_1,\ldots,y^{(n)}_L] 
$$
where
\begin{equation*}
	y^{(n)}_{\ell} \in \mathcal{Y}_{\ell} = \{1,\ldots,K_{\ell} \},
\end{equation*}
with $K_{\ell}\in \mathbb{N}_+$ being the finite number of values associated with the $\ell$-th label.
For some test instance $\xtest$, and provided that we know the conditional distribution $p(\y|\x)$, the MOC solution is given by
\begin{equation}
	\label{eq:LP}
	\ypred = h(\xtest) = \argmax_{\y \in {\boldsymbol \Y}} p(\y|\xtest).
\end{equation}
Once more, $p(\y|\x)$ is usually unknown and has to be estimated from the training data, $\D = \{(\x^{(n)},\y^{(n)})\}_{n=1}^N$, in order to construct the model $h$.
Therein precisely lies the main challenge behind MOC, since $h$ must select one out of $|\mathcal{Y}|=K^L$ possible values\footnote{A simplification of $K_1 \times K_2 \times \cdots \times K_L$ (to keep notation cleaner).}; clearly a much more difficult task than in \Eq{eq:MC}.
Furthermore, finding $\ypred$ for a given $\xtest$ and $p(\y|\x)$ is quite challenging from a computational point of view for large values of $K$ and $L$ \cite{MCC2,BCC}.

\begin{figure*}
\begin{center}
		 \includegraphics[scale=0.5]{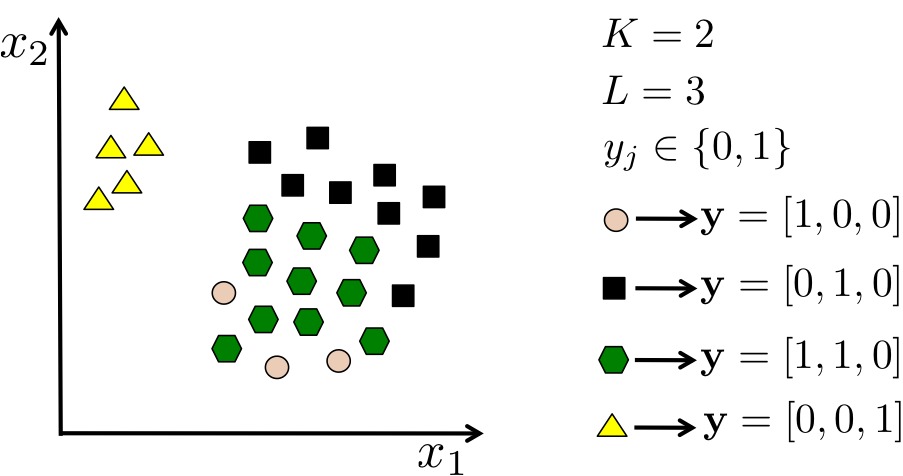}
	\end{center}
\caption{\label{fig:intro2} Toy example of multi-label classification (MLC), with $K=2$ possible values for each label and $L=3$ labels (thus $y_j \in \{0,1\}$ for $j=1,2,3$) and, implicitly, $D=2$ features. Circles, squares and triangles are elements with only one active label (i.e., either $y_1=1$ or $y_2=1$, but not both). Hexagons show vectors such that $y_1=y_2=1$. 
}
\end{figure*}

In MLC, all labels are binary labels, namely $K_{\ell}=2$ for $\ell=1,\ldots,L$, with the two possible label values typically notated as $y_{\ell} \in \{0,1\}$ or $y_{\ell} \in \{-1,+1\}$.
\Fig{fig:intro2} shows one toy example of MLC with three labels (thus $\y \in \{0,1\}^3$).
Because of the strong co-occurrence, we can interpret that the first label ($y_1=1$)  implies the second label ($y_2=1$) with high probability, but not the other way around.
When learning the model $h(\x)$ in \eqref{eq:LP}, the goal of MLC (and MOC in general) is capturing this kind of dependence among labels in order to improve classification performance; and to do this efficiently enough to scale up to the size of the data in the application domains of interest. This typically means connecting labels (i.e., learning labels together) in an appropriate structure. \Tab{TableNot} summarizes the main notation used in the paper.

\begin{table}[!hbt]
\small
\begin{center}
\caption{Summary of the main notation.}
\label{TableNot}
\begin{tabular}{ll}
\toprule
{\bf Notation } & {\bf Description}  \\
\midrule
	$\x = [x_1,\ \ldots,\ x_D]^{\top}$ & instance / input vector; $\x \in \mathbb{R}^D$\\ 
	$y \in \{0,1\}$ & a label (binary variable) \\
	$y \in \{1,\ldots,K\}$ & an output (multi-class variable), $K$ possible values \\
	$\y = [y_1,\ldots,y_L]^{\top}$ & $L$-dimensional label/output vector\\
	$\D = \{(\x^{(n)},\y^{(n)})\}_{n=1}^N$ & Training data set, $n=1,\ldots,N$\\
$\yp = h(\xtest)$ & binary or multi-class classification (test instance $\xtest$) \\
$\ypred = h(\xtest)$ & multi-label multi-output classification (MLC, MOC)\\
\bottomrule
\end{tabular}
\end{center}
\end{table}

\section{Background and Related Work}
\label{sec:prior}

In this section,  we step through some of the most relevant methods for MLC/MOC recently developed, as well as several works specifically related to the novel method, presented in later sections. 
All the methods discussed here, and also the CT method presented in \Sec{sec:CT}, aim to build a model for $h(\x)$ in \eqref{eq:LP} by first selecting a suitable model for the label joint posterior distribution $p(\y|\x)$ and then using this model to provide a prediction $\ypred$ to a new test input $\xtest$.  It is in the first step where state-of-the-art methods present a complexity bottleneck to deal with large sets of labels and where CT offers a significantly better complexity-performance trade-off. 



\begin{figure}
\begin{tabular}{cc}
\includegraphics[scale=0.8]{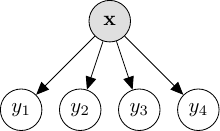} & \includegraphics[scale=0.8]{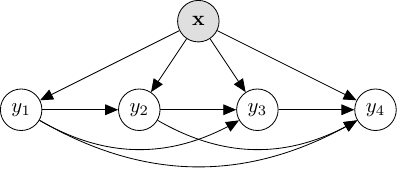} \\
 (a) Independent Classifiers (\texttt{IC}) & (b) Classifier Chain (\texttt{CC})\\\\
\includegraphics[scale=0.8]{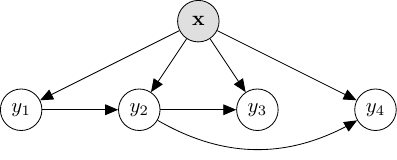} & \includegraphics[scale=0.8]{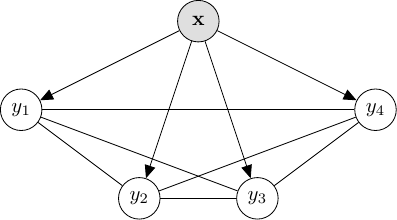}\\ 
(c)  Bayesian Classifier Chain (\texttt{BCC}) & (d) Conditional Dependency Network (\texttt{CDN})
\end{tabular}
\caption{\label{fig:models}Several multi-label methods depicted as directed/undirected graphical models.} 
\end{figure}

%
%
%

\subsection{Independent Classifiers}
A naive solution to multi-output learning is training $L$ $K$-class models as in \Eq{eq:MC}, i.e., $L$ \key{independent classifiers} (\texttt{IC}),\footnote{In the MLC literature, the \texttt{IC} approach is also known as the \key{binary relevance} method.} and using them to classify $L$ times a test instance $\xtest$,  as $[\yp_1,\ldots,\yp_L] = [h_1(\xtest),\ldots,h_L(\xtest)$].
\texttt{IC} is represented by the directed graphical model shown in \Fig{fig:models} (a).
Note that this approach implicitly assumes the independence among the target variables, i.e., $ p(\y|\x) \equiv \prod_{\ell=1}^L p(y_{\ell}|\x) $, which is not the case in most (if not all) multi-output datasets.

\subsection{Classifier Chains}
\label{seq:cc}

The \emph{classifier chains} methodology is based on the decomposition of the conditional probability of the label vector $\y$ using the product rule of probability:
\begin{align}
	p(\y|\x) &= p(y_1|\x) \prod_{\ell=2}^L p(y_{\ell}|y_1,\ldots,y_{\ell-1},\x)\label{eq:chain} \\
			 &\approx f_1(\x) \prod_{\ell=2}^L f_{\ell}(\x,y_1,\ldots,y_{\ell-1})\label{eq:approx_joint},
\end{align}
which is approximated with $L$ probabilistic classifiers, $f_{\ell}(\x,y_1,\ldots,y_{\ell-1})$
. As a graphical model, this approach is illustrated by \Fig{fig:models} (b).

The complexity associated to learn \Eq{eq:approx_joint} increases with $L$, but with a fast greedy inference, as in \cite{ECC2}, it reduces to 
\begin{equation}
	\hat{y}_{\ell} = \argmax_{y_{\ell}} p(y_{\ell}|y_1,\ldots,y_{\ell-1},\x)\texttt{,} \label{eq:CC}
\end{equation}
for $\ell=1,\ldots,L$. This is not significant for most datasets, and time complexity is close to that of \texttt{IC} in practice. In fact, it would be identical if not for the extra $y_1,\ldots,{y_\ell-1}$ attributes.

With greedy inference comes the concern of error propagation along the chain, since an incorrect estimate $\hat{y}_\ell$ will negatively affect all following labels. However, this problem is not always serious, and easily overcome with an ensemble \cite{ECC2,PCC}. Therefore, although there exist a number of approaches for avoiding error propagation via exhaustive iteration or various search options \cite{PCC, BeamSearch, MCC2}, we opt for the ensemble approach.  

\subsection{Bayesian Classifier Chains}
\label{sec:BCC}

Instead of considering a fully parameterized Markov chain model for $p(\y|\x)$, we can use a simpler Bayesian network. Hence, \eqref{eq:chain} becomes
\begin{align}\label{eq: BN}
p(\y|\x)=\prod_{\ell=1}^{L} p(y_{\ell}|\y_{\pa{\ell}},\x),
\end{align}
where $\pa{\ell}$ are the parents of the $\ell$-th label, as proposed in \cite{BCC,LEAD}, known as as Bayesian Classifier Chains (\texttt{BCC}), since it may remind us of Bayesian networks. Using a structure makes training the individual classifiers faster, since there are fewer inputs to them, and also speeds up any kind of inference. 
\Fig{fig:models} (c) shows one example of many possible such network structures.

Unfortunately, finding the optimal structure is NP hard due to an impossibly large search space.
Consequently, a recent point of interest has been finding a good suboptimal structure, such that \Eq{eq: BN} can be used.
The literature has focused around the idea of label dependence (see \cite{OnLabelDependence2} for an excellent discussion). 
The least complex approach is to measure \emph{marginal label dependence}, i.e., the relative co-occurrence frequencies of the labels. Such approach has been considered in \cite{BCC,BNFS}. In the latter, the authors exploited the \key{frequent sets} approach \cite{FreqSets}, which measures the co-occurrence of several labels, to incorporate edges into the Bayesian network.
However, they noted problems with attributes and negative co-occurrence (i.e., mutual exclusiveness; labels whose presence indicate the absence of others).
%
The resulting  algorithm (hereafter referred to as the \textsc{fs} algorithm) can deal with moderately large datasets, but the final network construction approach ends up being rather involved. 

%

Finding a graph based on \emph{conditional} label dependence is inherently more demanding, because the input feature space must be taken into account, i.e.,  classifiers must be trained. Of course, training time is a strongly limiting factor here. However, 
%
a particularly interesting approach to modelling conditional dependence, the so-called \textsc{lead} method, was presented in \cite{LEAD}.
This scheme tries to remove first the dependency of the labels on the feature set, which is the common parent of all the labels, to facilitate learning the label dependencies.
In order to do so, \textsc{lead} trains first an independent classifier for each label (i.e., it builds $L$ independent models, as in the \texttt{IC} approach), and then uses the dependency relations in the residual errors of these classifiers to learn a Bayesian network following some standard approach (the errors can in fact be treated exactly as if they were labels and plugged, e.g., into the \textsc{fs} approach).
\textsc{lead} is thus a fast method for finding conditional label dependencies, and has shown good performance on small-sized datasets.
%

Neither the \textsc{fs} nor the \textsc{lead} methods assume any particular constraint on the underlying graph and are well suited for MOC in the high dimensional regime because of their low complexity.
However, if  the underlying directed graph is sparse, the PC algorithm and its modifications \cite{KalischB07,YehezkelL09} are the state-of-the-art solution in directed structured learning.
%
%
%
The PC-algorithm runs in the worst case in exponential time (as a function of the number of nodes), but if the true underlying graph is sparse, this reduces to a polynomial runtime. However, this is typically not the case in MLC/MOC problems.
%

%
For the sake of comparison between the different MLC/MOC approaches, in this paper we only consider the \textsc{fs} and \textsc{lead} methods to infer direct dependencies between labels. In order to delve deeper into the issue of structure learning using the \textsc{fs} and \textsc{lead}  methods, in \ref{sec:appendixA} we have generated a synthetic dataset, where the underlying structure is known, and compare their solutions and the degree of similarity with respect to the true graphical model. As these experiments illustrate, one of the main problems behind learning the graphical model structure from scratch  is that we typically get too dense networks, where we cannot control the complexity associated to training and evaluating each one of the probabilistic classifiers corresponding to the resulting factorization in \eqref{eq: BN}. This issue is solved by the classifier trellis method proposed in \Sec{sec:CT}.

%

\subsection{Conditional Dependency Networks (\texttt{CDN})}
\label{sec:CDNG}

Conditional Dependency Networks  represent an alternative approach, in which the conditional distribution $p(\y|\x)$ factorizes according to an undirected graphical model, i.e.,
\begin{align}\label{eq:undirected}
p(\y|\x)=\frac{1}{Z}\prod_{q=1}^{Q}\phi_{q}(\y_q|\x),
\end{align}
where $Z$ is a normalizing constant, $\phi_{I}(\cdot)$ is a positive function or potential, and $\y_q$ is a subset of the labels (a \emph{clique} in the undirected graph).
%
%
The notion of directionality is dropped, thus simplifying the task of learning the graph structure.
Undirected graphical models are more natural for domains such as spatial or relational data.
Therefore, they are well suited for tasks such as image segmentation (e.g., \cite{Littman09}, \cite{Russell09}), and regular MLC problems (e.g., \cite{CollectiveMC}).

Unlike classifier chain methods, a \texttt{CDN} does not construct an approximation to $p(\y|\x)$ based on a product of probabilistic classifiers.
In contrast, for each conditional probability of the form
\begin{align}
p(y_{\ell}|y_1^{(t)},\ldots,y_{\ell-1}^{(t)},y_{\ell+1}^{(t-1)},\ldots,y_{L}^{(t-1)},{\bf x}),
\end{align}
a probabilistic classifier is learnt.
In an undirected graph, where all the labels that belong to the same clique are connected to each other, it is easy to check that
\begin{align}\label{eq:Markov}
p(y_{\ell}|y_1^{(t)},\ldots,y_{\ell-1}^{(t)},y_{\ell+1}^{(t-1)},\ldots,y_{L}^{(t-1)},{\bf x})=p(y_{\ell}|\y_{\ne{\ell}},{\bf x}),
\end{align}
where $\y_{\ne{\ell}}$ is the set of variables connected to $y_{\ell}$ in the undirected graph\footnote{In other words, $\y_{\ne{\ell}}$ is the so called \emph{Markov blanket} of $y_{\ell}$ \cite{Barber}.}. Finally, $p(y_{\ell}|\y_{\ne{\ell}},{\bf x})$ for $\ell=1,\ldots,L$ is approximated by a probabilistic classifier $f_{\ell}(\y_{\ne{\ell}},{\bf x})$.
%
%

In order to classify a new test input $\xtest$, approximate inference using Gibbs sampling is a viable option. In  \ref{sec:inference}, we present the formulation of  Monte Carlo approaches (including Gibbs sampling) specially tailored to perform approximate inference in MLC/MOC methods based on Bayesian networks and undirected graphical models.

\subsection{Other MLC/MOC Approaches}
\label{sec:other}

A final note on related work: there are many other `families' of methods designed for multi-label, multi-output and structured output prediction and classification, including many `algorithm adapted' methods.
A fairly complete and recent overview can be seen in \cite{Review} for example.
However, most of these methods suffer from similar challenges as the classifier chains family; and similarly attempt to model dependence yet remain tractable by using approximations and some form of randomness \cite{RAKEL,EPS,VensForest}. To cite a more recent example, \cite{RandomGraphEnsembles} uses `random graphs' in a way that resembles \cite{CDN}'s \texttt{CDN}, since it uses undirected graphical models, and  \cite{BCC}'s \texttt{BCC} in the sense of the randomness of the graphs considered.

\subsection{Comparison of State-of-the-art Methods for Extracting Structure}
\label{sec:toy}

It is our view that many methods employed for multi-label chain classifiers have not been properly compared in the literature, particularly with regard to their method for finding structure.
There is not yet any conclusive evidence that modelling the marginal dependencies (among $Y$) is enough, or whether it is advisable to model the conditional dependencies also (for best performance); and how much return one gets on a heavy investment in searching for a `good' graph structure, over random structures. 

We performed two experiments to get an idea; comparing the following methods:
\begin{center}
	\centering
	\begin{tabularx}{\textwidth}{lX}
		\texttt{IC}& Labels are independent, no structure.\\
		\texttt{ECC}& Ensemble of $10$ \texttt{CC}, each with random order \cite{ECC2}\\
		\texttt{EBCC}-\textsc{fs} & Ensemble of $10$ \texttt{BCC}s \cite{BCC}, based on marginal dependence \cite{BNFS}\\
		\texttt{EBCC}-\textsc{lead}& as above, but on conditional dependence, as per \cite{LEAD}\\
		\texttt{OCC}& the \emph{optimal} \texttt{CC} -- of all $L!$ possible (complete) chain orders\\
	\end{tabularx}
\end{center}

\begin{table}[t]
	\centering
	\caption{\label{tab:2}Comparison of the classification accuracy of existing methods with $5\times$ CV (Cross validation). We chose the two smallest datasets so that the `optimal' \texttt{OCC} could complete.}
	\footnotesize
	\begin{tabular}{rcccccc}
		\toprule
		& \texttt{IC}       & \texttt{ECC}       & \texttt{OCC} & \texttt{EBCC}-\textsc{fs}  & \texttt{EBCC}-\textsc{lead} \\
		\midrule                                                
		\textsf{Music}	& $0.517 \pm 0.03$  & $0.588 \pm 0.03$  & $0.580 \pm 0.04$                    & $0.582 \pm 0.02$    & $0.578 \pm 0.03$ \\
		\textsf{Scene}	& $0.595 \pm 0.02$  & $0.698 \pm 0.01$  & $0.699 \pm 0.01$                  &  $0.644 \pm 0.02$    & $0.645 \pm 0.03$ \\
		\bottomrule
	\end{tabular}
\end{table}

Results are shown in \Tab{tab:2} of $5\times$CV (Cross validation) on two small real-world datasets (\textsf{Music} and \textsf{Scene}); small to ensure a comparison with \texttt{OCC} ($L!=720$ possible chain orderings).

\begin{figure}[h]
	\centering
\subfloat[]{	\includegraphics[width=0.45\textwidth]{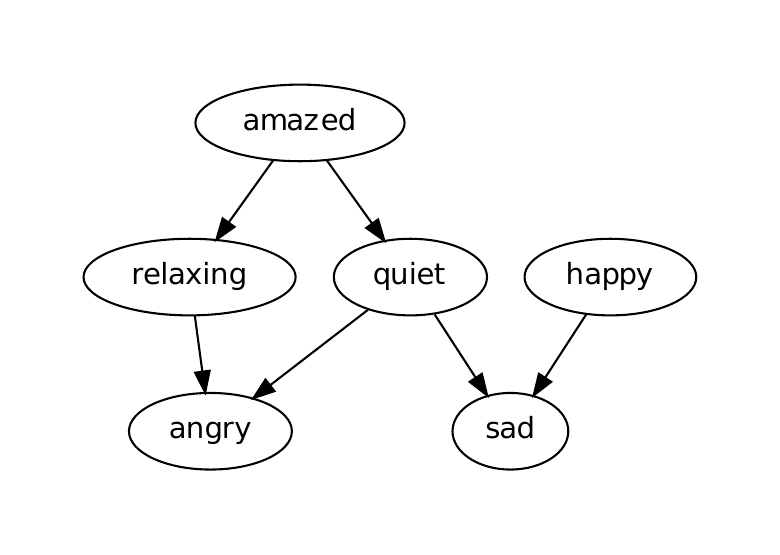}}
\subfloat[]{	\includegraphics[width=0.40\textwidth]{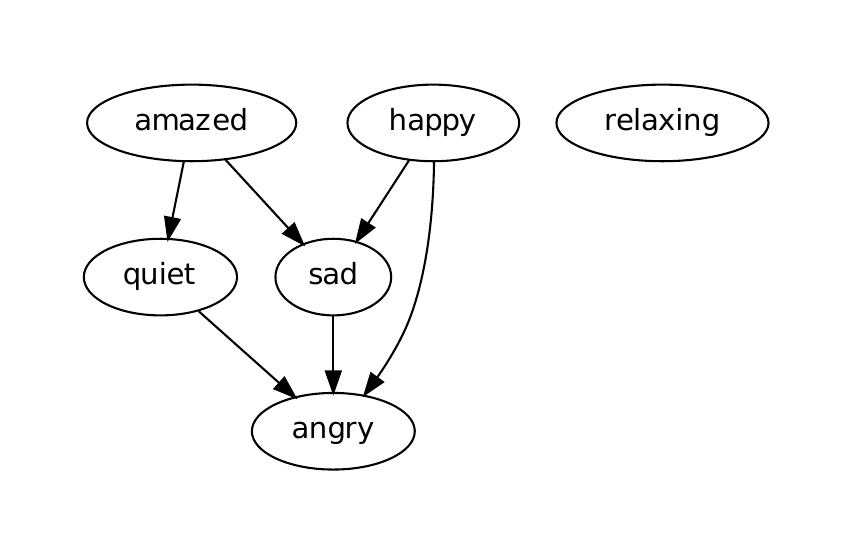}}
	\caption{\label{fig:musicrecon} Graphs derived from the \textsf{Music} dataset, with links based on (a) marginal dependence (FS, label-frequency) and (b) conditional dependence (LEAD, error-frequency).  Here we have based the links on mutual information, therefore links represent both co-occurrences (e.g., $\textsf{quiet} \rightarrow \textsf{sad}$) and mutual exclusiveness (e.g., $\textsf{happy} \rightarrow \textsf{sad}$).  Generally, we see that the graph makes intuitive sense; \textsf{amazed} and \textsf{happy} are neither strongly similar nor opposite emotions, and thus there is not much benefit in modelling them together; same with \textsf{angry} and \textsf{sad}.}
\end{figure}

\Fig{fig:musicrecon} shows an example of the structure found by \texttt{BCC}-\textsc{fs} and \texttt{BCC}-\textsc{lead} in a real dataset, namely \textsf{Music} (emotions associated with pieces of music).
As a base classifier, we use support vector machines (SVMs), fitted with logistic models (as according to \cite{SMOM}) in order to obtain a probabilistic output, with the default hyper-parameters provided in the SMO implementation of the Weka framework \cite{WEKA}.
\Tab{tab:2} confirms that \texttt{IC}'s assumption of independence harms its performance, and that the bulk of the MOC literature is justified in trying to overcome this.
However, it also suggests that investing factorial and exponential time to find the best-fitting chain order and label combination (respectively) does not guarantee the best results.
In fact, by comparing the results of \texttt{ECC} with \texttt{EBCC}-\textsc{lead} and \texttt{EBCC}-\textsc{fs}, even the relatively higher investment in conditional label dependence over marginal dependence (\texttt{EBCC}-\textsc{lead} vs \texttt{EBCC}-\textsc{fs}) does not necessarily pay off.
Finally, \texttt{ECC}'s performance is quite close to \texttt{MCC}.
Surprisingly, the \texttt{ECC} method tends to provide excellent performance, even though it only learns randomly ordered (albeit fully connected) chains.
However, as discussed in \Sec{seq:cc}, its complexity is prohibitively large in high dimensional MLC problems.

\begin{figure}
\centering
 \includegraphics[width=1\textwidth]{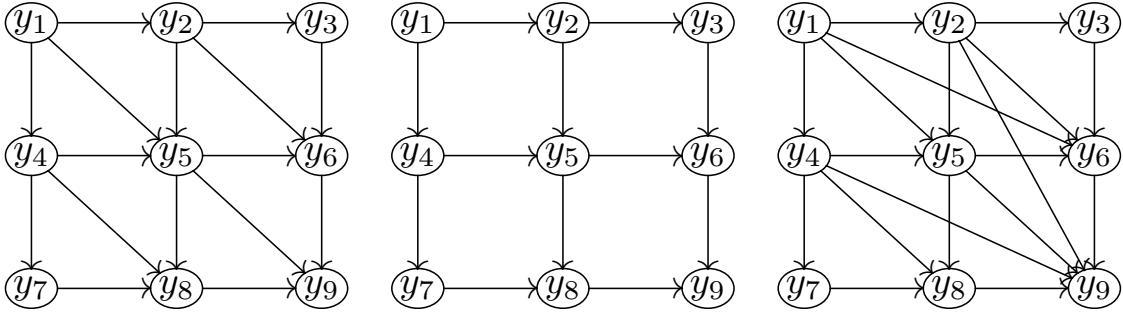}
 \caption{\label{fig:trellis}Three possible directed trellises for $L=9$. Each trellis is defined by a fixed pattern for the parents of each vertex (varying only near the edges, where parents are not possible). Note that no directed loops can exist.
 }
\end{figure}

\section{A scalable  Approach: Classifier Trellis (\texttt{CT}) } 
\label{sec:CT}

The goal for a highly scalable CC-based method brings up the common question: which structure to use.
On the one hand, ignoring important dependency relations will harm performance on typical MLC problems.
On the other hand, assumptions must be made to scale up to large scale problems.
Even though in certain types of MLC  problems there is a clear notion of the local structure underlying the labels (e.g., in image segmentation pixels next to each other should exhibit interdependence), this assumption is not valid for general MLC problems, where the $4$th and $27$th labels might be highly correlated for example.
Therefore, we cannot escape the need to discover structure, but we must do it efficiently.
Furthermore, the structure used should allow for fast inference. 

Our proposed solution is the \emph{classifier trellis} (\texttt{CT}).
To relieve the burden of specification `from scratch', we maintain a fixed structure, namely, a \emph{lattice} or \emph{trellis} (hence the name).
This escapes the high complexity of a complete structure learning (as in \cite{ECC2} and \cite{PCC}), and at the same time avoids the complexity involved in discovering a structure (e.g., \cite{BCC} and \cite{LEAD}). Instead, we a impose a structure a-priori, and only seek an improvement to the order of labels within that structure. 

\Fig{fig:trellis} gives three simple example trellises for $L=9$ (specifically, we use the first one in experiments). Each of the vertices $\ell \in \{1,\ldots,L\}$ of the trellis corresponds to one of the labels of the dataset. Note the relationship to \Eq{eq: BN}; we simply fix the same pattern to each $\pa{\ell}$. Namely, the parents of each label are the labels laying on the vertices above and to the left in the trellis structure (except, obviously, the vertices at the top and left of the trellis which form the border). A more linked structure will model dependence among more labels.

Hence, instead of trying to solve the NP hard structure discovery problem, we use a simple heuristic (label-frequency-based pairwise mutual information) to place the labels into a fixed structure (the trellis) in a sensible order: one that tries to maximize label dependence between parents and children. This ensures a good structure, which captures some of the main label dependencies, while maintaining scalability to a large number of labels and data. Namely, we employ an efficient \key{hill climbing} method to insert nodes into this trellis according to marginal dependence information, in a manner similar to the \textsc{fs} method in \cite{FreqSets}.

The process followed is outlined in \Code{code:CT}, to which we pass a pairwise matrix of mutual information, where
$$
I(Y_{\ell};Y_k)  = \sum_{y_{\ell} \in \Y_{\ell}}\sum_{y_k \in \Y_k} p(y_{\ell},y_k)
	\log \Big(\frac{p(y_{\ell},y_k)}{p(y_{\ell})p(y_k)}\Big).
$$
Essentially, new nodes are progressively added to the graph based on the mutual information.
Since the algorithm starts by placing a \emph{random} label in the upper left corner vertex, a different trellis will be discovered for different random seeds.
Each label $y_{\ell}$ is directly connected to a fixed number of parent labels in the directed graph (e.g., in Figure \ref{fig:trellis} each node has two parents in the first two graphs, and four in the third one) -- except for the border cases where no parents are possible.
\begin{code}[t]
    \caption{\label{code:CT} Constructing a Classifier Trellis}
\removelatexerror
\begin{algorithm}[H]
    \SetKwInOut{Input}{input}
    \SetKwInOut{Output}{output}
    \Input{
        $\mat{Y}$ (an $L \times N$ matrix of labels),
        $W$ (width, default: $\sqrt{L}$),
        a parent-node function $\pa{\cdot}$
    }
    \Begin{
        $\mat{I} \gets \textsc{learnDependencyMatrix}(\mat{Y})$ \;
        $S = \textsc{shuffle}(\{1,\ldots,L\})$ \;
		$s_1 = S_1$\;
        \For{$\ell=2,\ldots,L$;}{
        $S = S \setminus s_{\ell-1}$\;
		$s_{\ell} = \argmax_{k \in S} \sum_{j \in \textsf{pa}(\ell)} I(y_j;y_k)$\;
        }
    }
    \Output{
        the trellis structure, $\begin{bmatrix}
					s_1 & s_2 & \cdots & s_{W-1} & s_W\\
				 \vdots & \vdots & \ddots & \vdots & \vdots \\
				 s_{L-W} & s_{L-W+1} & \cdots & s_{L-1} & s_{L} \\
		\end{bmatrix}$.}          
\end{algorithm}
    We henceforth assume that $y_{\ell} = y_{s_{\ell}}$.
\end{code}
The computational cost of this algorithm is $O(L^2)$, but due to the simple calculations involved, in practice it is easily able to scale up to tens of thousands of labels.
Indeed, we show later in \Sec{sec:experiments} that this method is fast and effective.
Furthermore, it is possible to limit the complexity by searching for some number $<L$ of labels (e.g., building clusters of labels).





Given a proper user-defined parent pattern $\pa{\ell}$ (see \Fig{fig:trellis}), we are ensured that the trellis obtained in \Code{code:CT} is a directed acyclic graph.
Hence, there is no need to check for cycles during construction, which is a very time consuming stage in many algorithms (e.g., in the PC algorithm, see \Sec{sec:BCC}).
We can now employ probabilistic classifiers to construct an approximation to $p(\y|\x)$ according to the directed graph. This approach is simply referred to as the \key{Classifier Trellis} (\texttt{CT}).
Afterwards, we can either do inference greedily or via Monte Carlo sampling (see \ref{sec:inference} for a detailed discussion of Monte Carlo methods). 

Alternatively, note that we can interpret the trellis structure provided by \Code{code:CT} in terms of an undirected graph,  following the approach of Classifier Dependency Networks, described in \Sec{sec:CDNG}.
For example, in \Fig{fig:trellis} (middle), we would get (with directionality removed) $\ne{5} = \{2,4,6,8\}$.
\begin{code}[t]
	\caption{\label{code:CT2} Classifier Trellis (\texttt{CT})}
	\removelatexerror
	\begin{algorithm}[H]
    \textsc{Train$(\D)$}\:
    \Begin{
	Find the directed trellis graph using \Code{code:CT}.
	
	\vspace{0.1cm}
	Train classifiers $f_1,\ldots,f_L$, each taking \emph{all parents} in the directed graph as additional features, such that
			$$
			f_{\ell}(\x) \approx p(y_{\ell}|y_{\pa{\ell}},\x).
			$$
		\vspace{-0.5cm}
    }
    \textsc{Test$(\xtest)$}\:
    \Begin{
        \For{$\ell=\{s_1,\ldots,s_L\}$;}{
			$ \yp_{\ell} = \argmax_{y_\ell} p(y_{\ell}|\y_{\pa{\ell}},\xtest) $\
        }
        
		\Return $\ypred = [\yp_1,\ldots,\yp_L]$
    }
\end{algorithm}

\end{code}
This compares\footnote{We did not add the diagonals in the set $\ne{5}$, since we wish that it be comparable in terms of the \emph{number} of connections} to $\pa{5} = \{1,2,4\}$.
We refer to this approach as the \key{Classifier Dependency Trellis} (\texttt{CDT}).

Both \texttt{CT} and \texttt{CDT} are outlined in \Code{code:CT2} and \Code{code:CDNG} respectively.
Some may argue that the undirected version is more powerful, since learning an undirected graph is typically easier than learning a directed graph that encodes causal relations.
However, \texttt{CDT} constructs an undirected graphical model where greedy inference cannot be implemented and we have to rely on (slower) Monte Carlo sampling methods in the test stage. This effect can be clearly noticed in \Tab{tab:times}. 
%
%

Finally, we will consider a simple ensemble method for \texttt{CT}, similar to that proposed in \cite{ECC} or \cite{BCC} to improve classifier chain methods: $M$ \texttt{CT} classifiers, each built from a different random seed, where the final label decision is made by majority voting. This makes the training time and inference $M$ times larger. We denote this method as \texttt{ECT}. A similar approach could be followed for the \texttt{CDT} (thus building an \texttt{ECDT}), but, given the higher computational cost of \texttt{CDT} during the test stage, we have concerns regarding the scalability of this approach, so we have excluded it from the simulations.

\begin{code}
	\caption{\label{code:CDNG} Classifier Dependency Trellis (\texttt{CDT})}
	\removelatexerror
	\begin{algorithm}[H]
    \textsc{Train$(\D)$}\:
    \Begin{
	Find the undirected trellis graph using \Code{code:CT}.
	
	Train classifiers $f_1,\ldots,f_L$, each taking \emph{all neighbouring labels} as additional features, such that
			$$
				f_{\ell}(\x) \approx p(y_{\ell}|\y_\ne{\ell},\x).
			$$
    }
    \textsc{Test$(\xtest)$}\:
    \Begin{
    \For{$t=1,\ldots,T_c,\ldots,T$;}{
        \For{$\ell=\textsc{shuffle}\{1,\ldots,L\}$;}{
				$ y^{(t)}_{\ell} \sim p(y_{\ell}|\y_\ne{\ell},\xtest) $\
        }
        
    }
    \Return $\ypred = \frac{1}{T-T_c} \big[\sum_{T_c < t < T}  y^{(t)}_1, \ldots, \sum_{T_c < t < T}  y^{(t)}_L\big]$
    }
		where $T_c$, $T$: settling time (discarded samples) and total iterations.
		\vspace{-0.2cm}
\end{algorithm}
\end{code}

\section{Experiments}
\label{sec:experiments}

\begin{table*}
\centering
\caption{\label{tab:methods} Methods tested in experiments.}
\footnotesize
\begin{tabularx}{\textwidth}{lXl}
	\toprule
	Key     & Description/Name & Reference \\ 
	\midrule
	\texttt{IC} & Independent Classifiers & Sec.\ \ref{sec:Introduction} \\
	\texttt{ECC} & Ensemble of $10$ random \texttt{CC}s   (majority vote per label)                & Sec.\ \ref{sec:prior}, \cite{ECC2} \\
	\texttt{MCC} & Best of $10$ random \texttt{CC} &  Sec.\ \ref{sec:prior}, \cite{MCC2} \\
	\texttt{EBCC} & Ensemble of $\min(10,L)$ \texttt{BCC}s (discovered directed graph) & Sec.\ \ref{sec:prior}, \cite{BCC} \\
	\texttt{CT} & Classifier trellis, as in \Fig{fig:trellis}, left (directed trellis) & Sec.\ \ref{sec:CT}, Alg.\ \ref{code:CT2}  \\
	\texttt{ECT} & Ensemble of $10$ \texttt{CT}s &  Sec.\ \ref{sec:CT}  \\
	\texttt{CDT} & Classifier dependency trellis, as in \Fig{fig:trellis}, middle (but \emph{undirected}); $T=100,T_c=10$ & Sec.\ \ref{sec:CT}, Alg.\ \ref{code:CDNG}\\
	\bottomrule
\end{tabularx}
\end{table*}

Firstly, in \Sec{sec:experiments1}, we compare \texttt{E}/\texttt{CT} and \texttt{CDT} with some high-performance MLC methods (namely \texttt{ECC}, \texttt{BCC}, \texttt{MCC}) that were discussed in \Sec{sec:prior}. We show that an imposed trellis structure can compete  with fully-cascaded chains such as \texttt{ECC} and \texttt{MCC}, and discovered structures like those provided by \texttt{BCC}.  Our approach based on trellis structures achieves similar MLC performance (or better in many cases) while presenting improved scalable properties and, consequently, significantly lower running times.

%
%
All the methods considered are listed in \Tab{tab:methods}. In \Tab{tab:complexity} we summarize their complexity. $DN$ represents the input dimensions (the dataset-dependent number of features times number of instances); we use $M=10$ ensemble methods and $T=100$ Gibbs iterations for \texttt{CDT}. While this complexity is just an intuitive measure, the experimental results reported in \Sec{sec:experiments1} confirm that \texttt{CT} running times are indeed very close to \texttt{IC}.

\Tab{tab:datasets} summarizes the collection of datasets that we use, of varied type and dimensions; most of them familiar to the MLC/MOC community \cite{RAKEL, PCC, ECC2}.

\begin{table}[th]
\caption{\label{tab:complexity} Complexity per algorithm, roughly sorted by training complexity. $DN$ represents the input dimensions (number of features times number of instances). Train complexity indicates roughly how many values are looked at by a classifier. Test complexity indicates how many (times) individual models are addressed as inference.}
\centering
\begin{tabular}{lll}
	\toprule
	method & train complexity & test complexity \\
	\midrule
	\texttt{IC}	& $O(LDN)$ & $O(L)$ \\
	\texttt{CT} & $O(LDN)$ & $O(L)$ \\
	\texttt{EBCC} & $O(M\times LDN)$  & $O(ML)$ \\
	\texttt{CDT} & $O(LDN)$ & $O(TL)$ \\
	\texttt{ECT} & $O(M\times LDN)$ & $O(ML)$ \\
	\texttt{ECC} & $O(M\times L[D+L]N)$ & $O(ML)$  \\
	\bottomrule
\end{tabular}
\end{table}
The last two sets (Local400 and Local10k) are synthetically generated and they correspond to the localization problem described in \Sec{sec:experiments2}.


%

We use some standard metrics from the multi-label literature (see, e.g., \cite{ExtML}), namely,

$$
	\textsc{Hamming Score} := \frac{1}{NL}\sum_{n=1}^{N}\sum_{j=1}^{L}{ \big[ y_{j}^{(n)} = \hat{y}_{j}^{(n)} \big] }\text{,}
$$
$$
	\textsc{Exact Match} := \frac{1}{N}\sum_{n=1}^{N}{ \big[ \y^{(n)} = \ypred^{(n)} \big]}\text{,}
$$
$$
\textsc{Accuracy}\footnote{Known commonly as \emph{Jaccard Index} in information retrieval circles.} := \frac{1}{N}\sum_{i=1}^{N} \frac{| \vec{y}^{(n)} \wedge \vec{\hat{y}}^{(n)} |}{| \vec{y}^{(n)} \vee \vec{\hat{y}}^{(n)}|},
$$
where $\big[ A \big]$ is an identity function, returning $1$ if condition $A$ is true, whereas $\wedge$ and $\vee$ are the bitwise logical AND and OR operations, respectively.

\begin{table}[t]
    \centering 
    \caption{\label{tab:datasets}A collection of datasets and associated statistics, where \texttt{LC} is \emph{label cardinality}: the average number of labels relevant to each example. }
	\begin{tabular}{rrrrcccr}
	\toprule
	    		& $N$  &$L$ &	$D$		&\texttt{LC} &Type 		 \\           
	\midrule           
	\data{Music}       &593    &6     &72      &1.87          &audio   \\    
	\data{Scene}   	&2407	&6     &294		&1.07          &image	 \\     
	\data{Yeast}   	&2417 	&14   	&103	   	&4.24          &biology \\     
	\data{Medical}		&978 	&45      &1449    &1.25          &medical/text	 \\     
	\data{Enron}   	&1702	&53    &1001    &3.38          &text	 \\     
	\data{TMC2007}    	&28596  &22    &500     &2.16 &text	 \\     
     \data{MediaMill}	&43907  &101    &120		&4.38	       &video	 \\     
    \data{Delicious}	&16105  &983    &500     &19.02         &text	 \\     
	\midrule
 	\data{Local400}	&10000  &400    & 30 &12.84         &localization     \\     
    \data{Local10k}	&50000  & 10000 & 30 &25.78         &localization    	 \\     
	\bottomrule
	\end{tabular}
\end{table}

\subsection{Comparison of \texttt{E}/\texttt{CT} to other MLC methods} 
\label{sec:experiments1}

First of all, to confirm that our proposed hill climbing strategy can actually have a beneficial effect (i.e., an improvement over a random trellis), we do a $10 \times 10$ cross validation (CV) on the smaller datasets.
Results are displayed in \Tab{tab:ECTcompare}.
A significant increase in performance can be seen, with a  decrease in the standard deviation (especially relevant in the Scene dataset).
This confirms that the proposed hill climbing strategy can help in optimizing performance and decreasing the sensitivity of \texttt{CT} wrt the initialization.

Then, we compare all the methods listed in Table \ref{tab:methods} on all the datasets of Table \ref{tab:datasets}. 
The results of predictive performance are displayed in Table \ref{tab:results}, and running times can be seen in Table \ref{tab:times}.
On the small datasets ($L<100$) we use Support Vector Machines as base classifiers, with fitted logistic models (as in \cite{SMOM}) for \texttt{CDN} (which requires probabilistic output for inference).
As an alternative, other authors have used Logistic Regression directly (e.g., \cite{PCC,CDN}) due to its probabilistic output.
In our experience, we obtain better and faster all-round performance with SVMs. 
Note that, for best accuracy, it is highly recommended to tune the base classifier. However, we wish to avoid this ``dimension'' and instead focus on the multi-label methods.
On the larger datasets (where $L > 100$) we instead use Stochastic Gradient Descent (SGD), with a maximum of only $100$ epochs, to deal with the scale presented by these large problems.
All our methods are implemented and made available within the Meka framework;\footnote{\url{http://meka.sourceforge.net}} an open-source multi-output learning framework based on the Weka machine learning framework \cite{WEKA}. The SMO SVM and SGD implementations pertain to Weka.

\begin{table}[t]
    \caption{\label{tab:ECTcompare} Comparing \texttt{CT} accuracy (on \textsf{Scene}, \textsf{Music}) without (just a random trellis) and with our `hill-climbing' (HC) method, over $10\times10$ CV.}
    \centering
	\begin{tabular}{rcc}
		\toprule
        & \texttt{CT}-random    & \texttt{CT}-HC \\
		\midrule                                                
        \textsf{Music}&$0.536 \pm 0.042$ &  \bf $0.541 \pm 0.031$ \\
        \textsf{Scene} &$0.639 \pm 0.210$	& \bf  $0.678 \pm 0.028$  \\
		\bottomrule
	\end{tabular}
\end{table}
Results confirm that both \texttt{ECT} and \texttt{CT} are very competitive in terms of performance and running time. Using the Hamming score as figure of merit and given the running times reported, \texttt{CT} is clearly superior to the rest of methods. Note that \Tab{tab:times} shows that \texttt{CT}'s running times are close  to \texttt{IC}, namely the method that neglects all statistical dependency between labels. With respect to exact match and accuracy performance results, which are measures oriented to the recovery of the whole set of labels, \texttt{ECT}  and \texttt{CT} achieve very competitive results with respect to \texttt{ECC} and \texttt{MCC}, which are high-complexity methods that model the full-chain of labels. \Tab{tab:times} reports training and test average times computed for the different MLC methods. We also include explicitly the number $L$ of labels per dataset. Note also that even though \texttt{ECT} considers a set of 10 possible random initializations, it does not significantly improve the performance of \texttt{CT} (a single initialization) for most cases, which suggest that the hill climbing strategy makes the \texttt{CT} algorithm quite robust with respect to initialization. Regarding scalability, note that the computed train/running times  for \texttt{CT} scale roughly linearly with the number of labels $L$. For instance, in the  \data{MediaMill} dataset $L=101$ while in \data{Delicious} $L$ is approximately one order of magnitude higher, $L=983$. As we can observe, \texttt{CT} running times  are multiplied approximately by a factor of 10 between both datasets. The same conclusions can be drawn also for the largest datasets, compare for instance running times between \data{Local400} and \data{Local10k}. Finally, \texttt{CDT}, our scalable modification of \texttt{CDN},  shows worse performance than \texttt{CT} while it requires larger test running times.

\begin{table*}
	\centering
	\footnotesize
	\caption{\label{tab:results}Predictive performance and dataset-wise (rank). DNF = Did Not Finish (within 24 hours or 2 GB memory).}

    \begin{center}
        Accuracy
    \end{center}
    \begin{tabular}{lrrrrrrr}
\hline
Dataset   & \alg{IC}&\alg{ECC}&\alg{MCC}&\alg{EBCC}&\alg{CT}&\alg{ECT}&\alg{CDT} \\
\hline
Music                &  0.483 (7) &  0.572 (2) &  0.568 (4) &  0.566 (5) &  0.577 (1) &  0.571 (3) &  0.505 (6)  \\ 
Scene                &  0.571 (7) &  0.684 (2) &  0.685 (1) &  0.618 (4) &  0.602 (6) &  0.666 (3) &  0.604 (5)  \\ 
Yeast                &  0.502 (6) &  0.538 (2) &  0.534 (4) &  0.535 (3) &  0.533 (5) &  0.541 (1) &  0.438 (7)  \\ 
Medical              &  0.699 (7) &  0.733 (3) &  0.721 (5) &  0.731 (4) &  0.755 (2) &  0.769 (1) &  0.704 (6)  \\ 
Enron                &  0.406 (5) &  0.448 (1) &  0.403 (6) &  0.441 (3) &  0.409 (4) &  0.443 (2) &  0.310 (7)  \\ 
TMC07                &  0.614 (5) &  0.645 (1) &  0.619 (4) &  0.628 (3) &  0.613 (6) &  0.633 (2) &  0.601 (7)  \\ 
MediaMill            &  0.379 (2) &  0.350 (5) &  0.349 (6) &  0.375 (3) &  0.391 (1) &  0.344 (7) &  0.374 (4)  \\ 
Delicious            &  0.122 (3) &    DNF     &  0.121 (5) &    DNF     &  0.127 (2) &  0.157 (1) &  0.122 (3)  \\ 
Local400       &  0.536 (7) &  0.625 (1) &  0.583 (3) &  0.578 (4) &  0.542 (6) &  0.587 (2) &  0.559 (5)  \\ 
Local10k         &  0.125 (4) &    DNF     &    DNF     &  0.175 (1) &  0.133 (3) &  0.166 (2) &  0.122 (5)  \\ 
\hline
avg rank  &  5.30      &  2.12      &  4.22      &  3.33      &  3.60      &  2.40      &  5.50      \\
\hline
\end{tabular}

    \begin{center}
        Hamming Score
    \end{center}
    \begin{tabular}{lrrrrrrr}
\hline
Dataset   & \alg{IC}&\alg{ECC}&\alg{MCC}&\alg{EBCC}&\alg{CT}&\alg{ECT}&\alg{CDT} \\
\hline
Music                &  0.785 (6) &  0.795 (3) &  0.789 (5) &  0.800 (1) &  0.798 (2) &  0.795 (3) &  0.768 (7)  \\ 
Scene                &  0.886 (4) &  0.892 (1) &  0.892 (1) &  0.886 (4) &  0.884 (6) &  0.891 (3) &  0.871 (7)  \\ 
Yeast                &  0.800 (1) &  0.789 (4) &  0.794 (2) &  0.787 (5) &  0.791 (3) &  0.786 (6) &  0.719 (7)  \\ 
Medical              &  0.988 (4) &  0.988 (4) &  0.989 (2) &  0.988 (4) &  0.990 (1) &  0.989 (2) &  0.986 (7)  \\ 
Enron                &  0.943 (1) &  0.940 (4) &  0.942 (3) &  0.939 (5) &  0.943 (1) &  0.939 (5) &  0.922 (7)  \\ 
TMC07                &  0.947 (2) &  0.948 (1) &  0.947 (2) &  0.946 (5) &  0.947 (2) &  0.946 (5) &  0.937 (7)  \\ 
MediaMill            &  0.965 (2) &  0.947 (7) &  0.958 (4) &  0.954 (5) &  0.966 (1) &  0.951 (6) &  0.965 (2)  \\ 
Delicious            &  0.982 (1) &    DNF     &  0.981 (4) &    DNF     &  0.982 (1) &  0.981 (4) &  0.982 (1)  \\ 
Local400       &  0.968 (3) &  0.969 (1) &  0.967 (6) &  0.968 (3) &  0.969 (1) &  0.968 (3) &  0.962 (7)  \\ 
Local10k         &  0.968 (1) &    DNF     &    DNF     &  0.968 (1) &  0.968 (1) &  0.968 (1) &  0.968 (1)  \\ 
\hline
avg rank  &  2.50      &  3.12      &  3.22      &  3.67      &  1.90      &  3.80      &  5.30      \\
\hline
\end{tabular}

    \begin{center}
        Exact Match
    \end{center}
    \begin{tabular}{lrrrrrrr}
\hline
Dataset   & \alg{IC}&\alg{ECC}&\alg{MCC}&\alg{EBCC}&\alg{CT}&\alg{ECT}&\alg{CDT} \\
\hline
Music                &  0.252 (7) &  0.327 (1) &  0.292 (5) &  0.302 (4) &  0.312 (2) &  0.312 (2) &  0.257 (6)  \\ 
Scene                &  0.491 (7) &  0.579 (2) &  0.638 (1) &  0.516 (5) &  0.542 (4) &  0.557 (3) &  0.503 (6)  \\ 
Yeast                &  0.160 (5) &  0.190 (3) &  0.212 (1) &  0.150 (6) &  0.198 (2) &  0.169 (4) &  0.067 (7)  \\ 
Medical              &  0.614 (4) &  0.612 (5) &  0.634 (3) &  0.612 (5) &  0.670 (1) &  0.655 (2) &  0.598 (7)  \\ 
Enron                &  0.121 (3) &  0.112 (5) &  0.126 (1) &  0.114 (4) &  0.123 (2) &  0.112 (5) &  0.067 (7)  \\ 
TMC07                &  0.330 (4) &  0.342 (2) &  0.345 (1) &  0.316 (6) &  0.341 (3) &  0.317 (5) &  0.263 (7)  \\ 
MediaMill            &  0.055 (2) &  0.034 (5) &  0.053 (3) &  0.019 (6) &  0.058 (1) &  0.007 (7) &  0.052 (4)  \\ 
Delicious            &  0.003 (3) &    DNF     &  0.006 (1) &    DNF     &  0.004 (2) &  0.002 (5) &  0.003 (3)  \\ 
Local400       &  0.064 (4) &  0.108 (2) &  0.129 (1) &  0.059 (6) &  0.079 (3) &  0.063 (5) &  0.029 (7)  \\ 
Local10k         &  0.000 (1) &    DNF     &    DNF     &  0.000 (1) &  0.000 (1) &  0.000 (1) &  0.000 (1)  \\ 
\hline
avg rank  &  4.00      &  3.12      &  1.89      &  4.78      &  2.10      &  3.90      &  5.50      \\
\hline
\end{tabular}



\end{table*}

\begin{figure}[h]
	\centering
	\subfloat[test][Accuracy]{
		\includegraphics[width=0.3\textwidth]{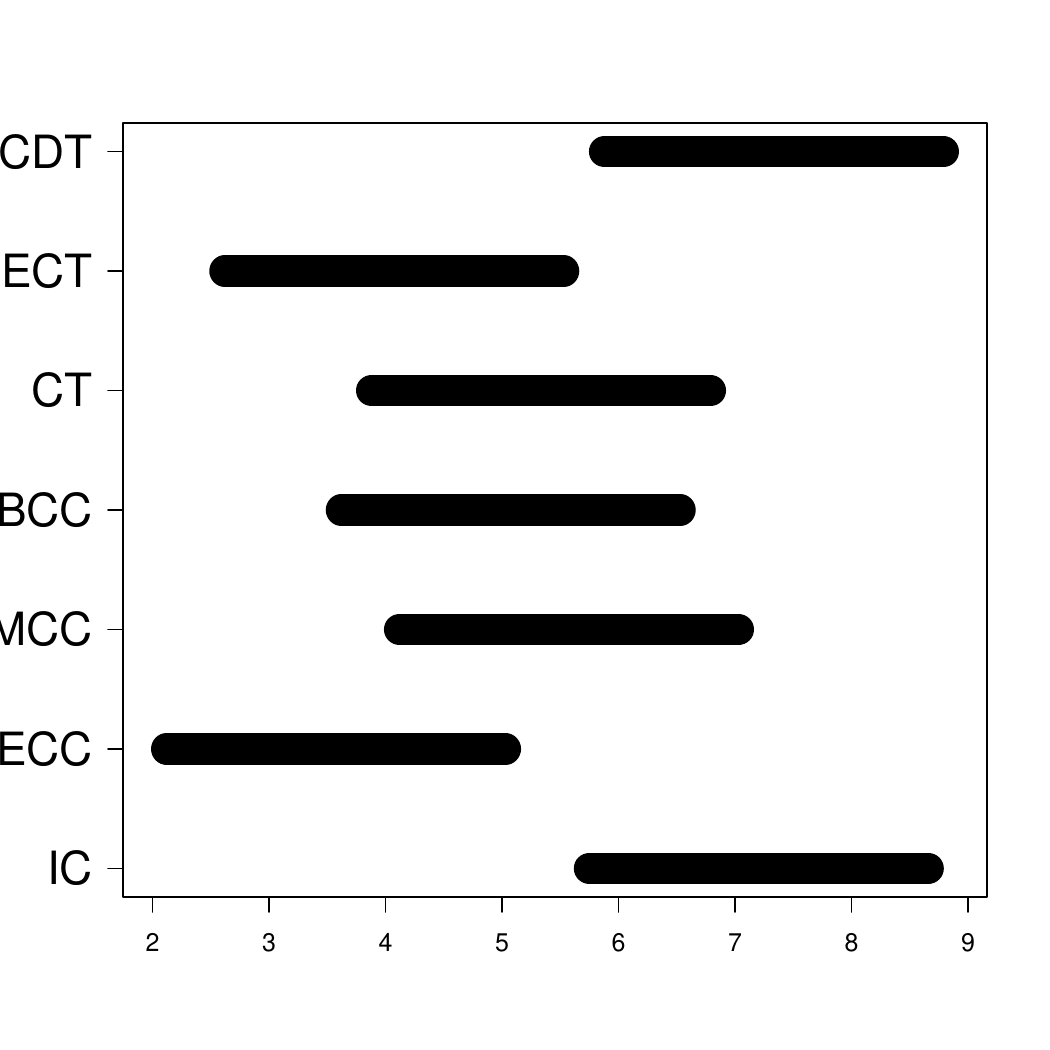}
	}
	\subfloat[test][Exact Match]{
		\includegraphics[width=0.3\textwidth]{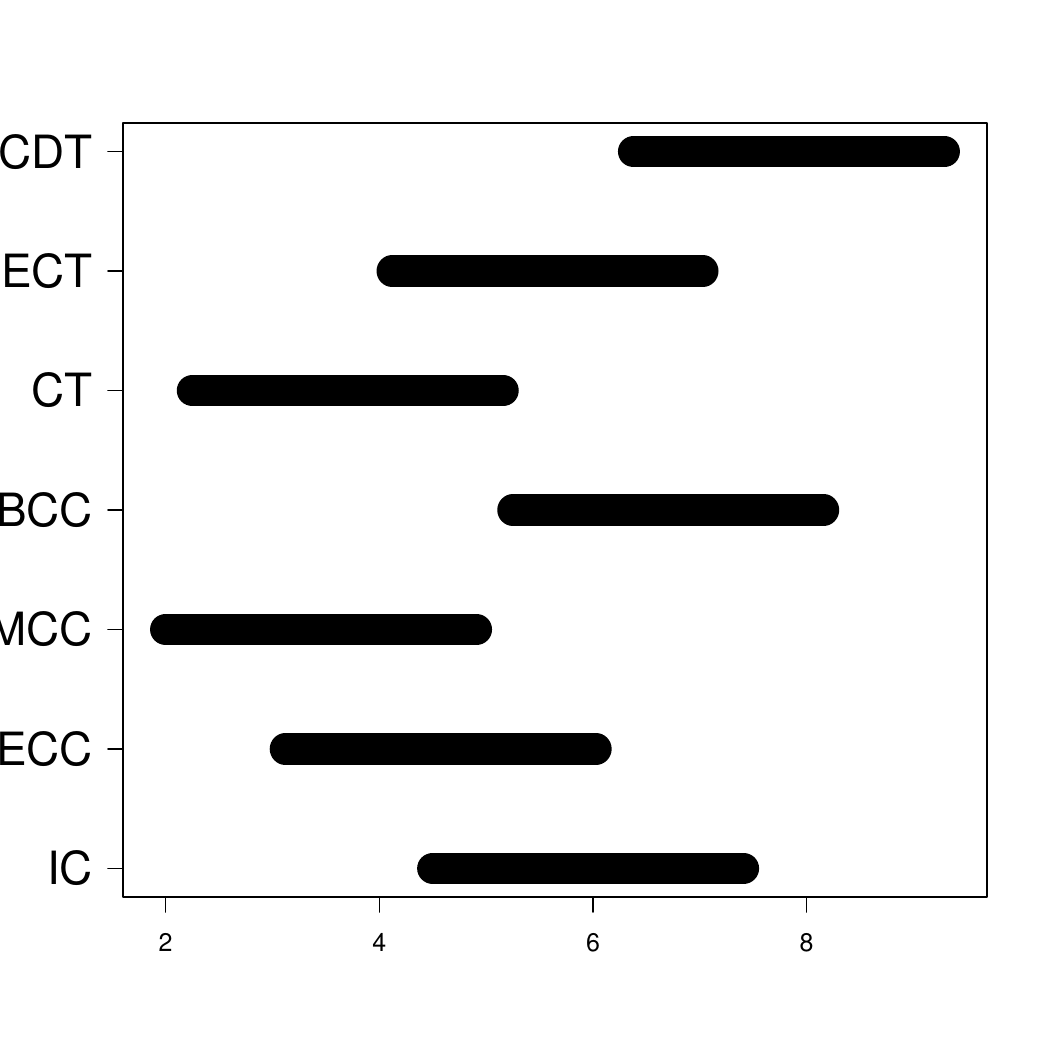}
	}
	\subfloat[test][Hamming Score]{
		\includegraphics[width=0.3\textwidth]{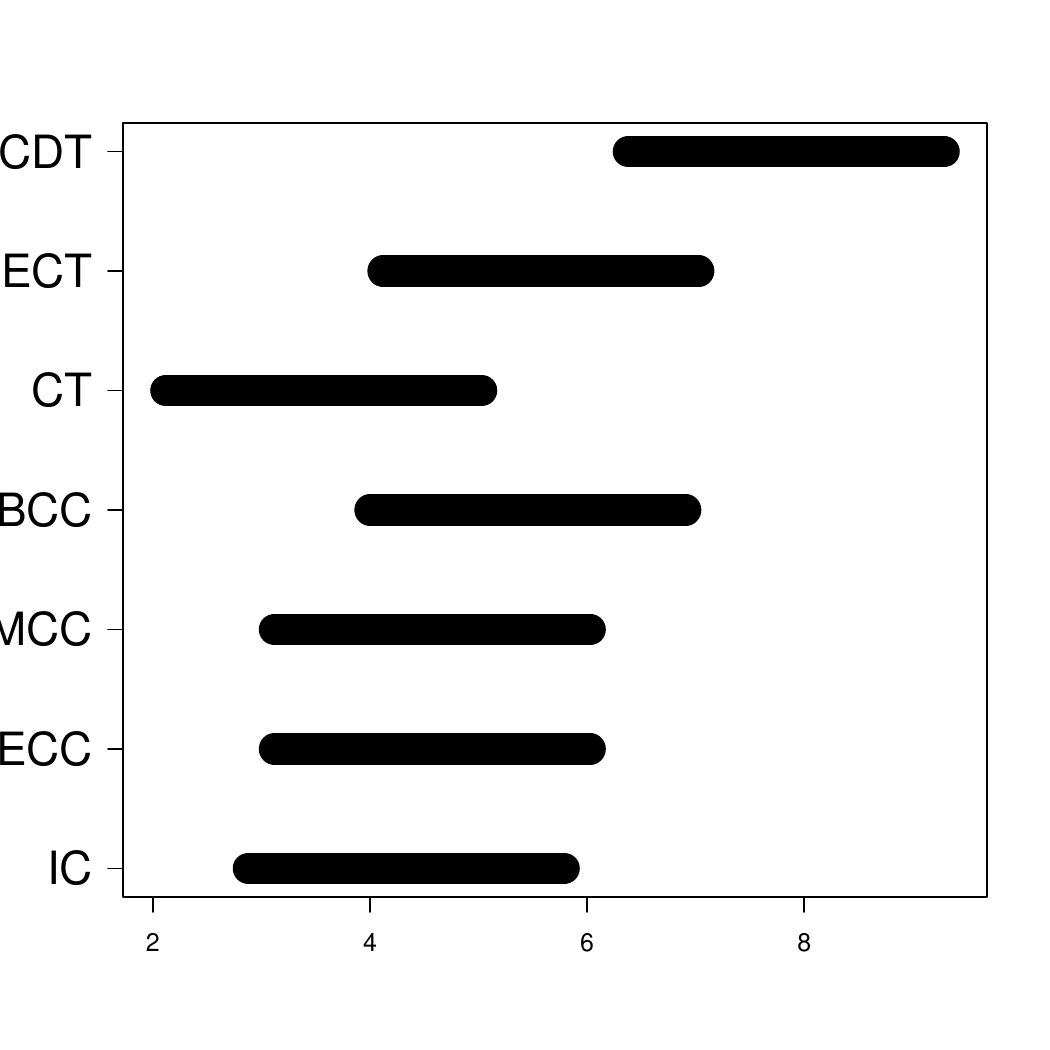}
	}\\
	\subfloat[test][Training Time]{
		\includegraphics[width=0.3\textwidth]{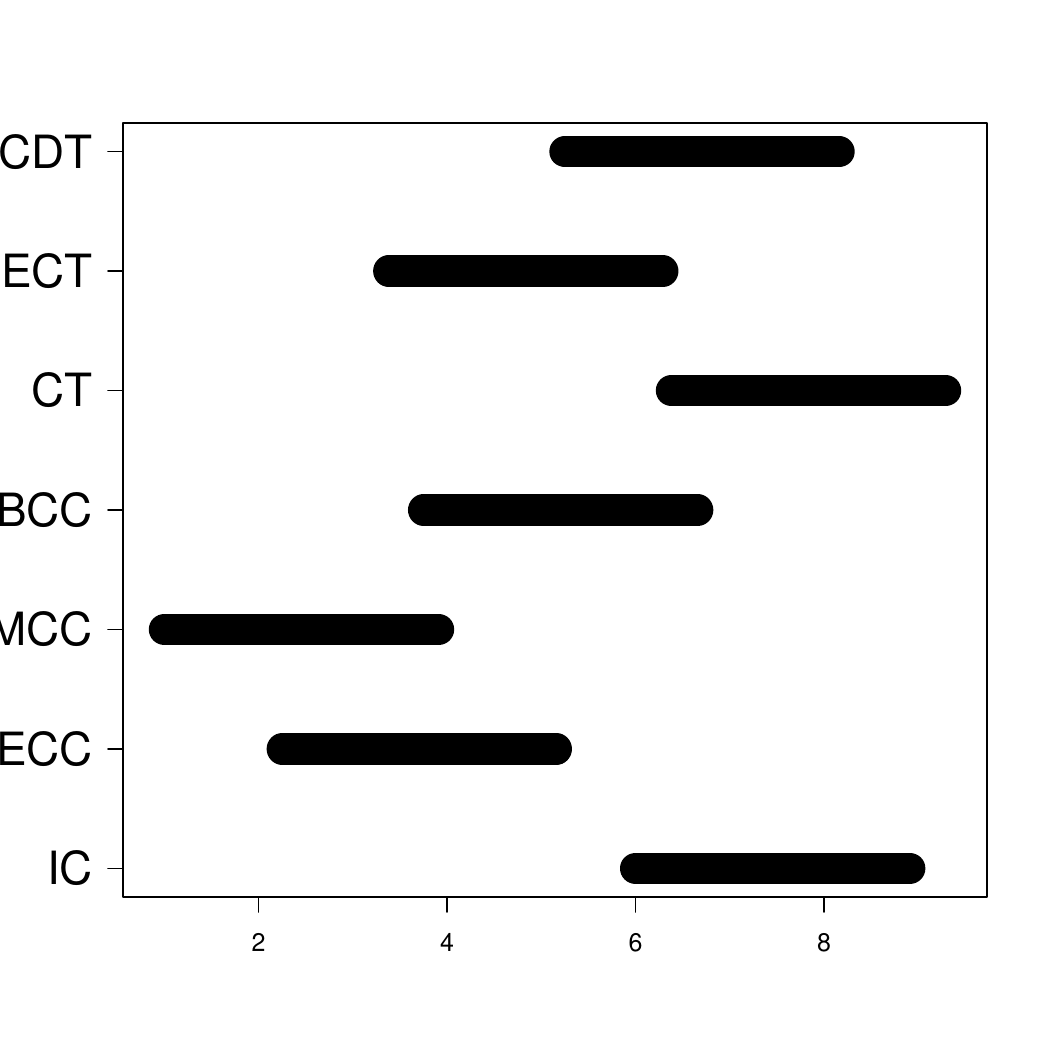}
	}
	\subfloat[test][Test Time]{
		\includegraphics[width=0.3\textwidth]{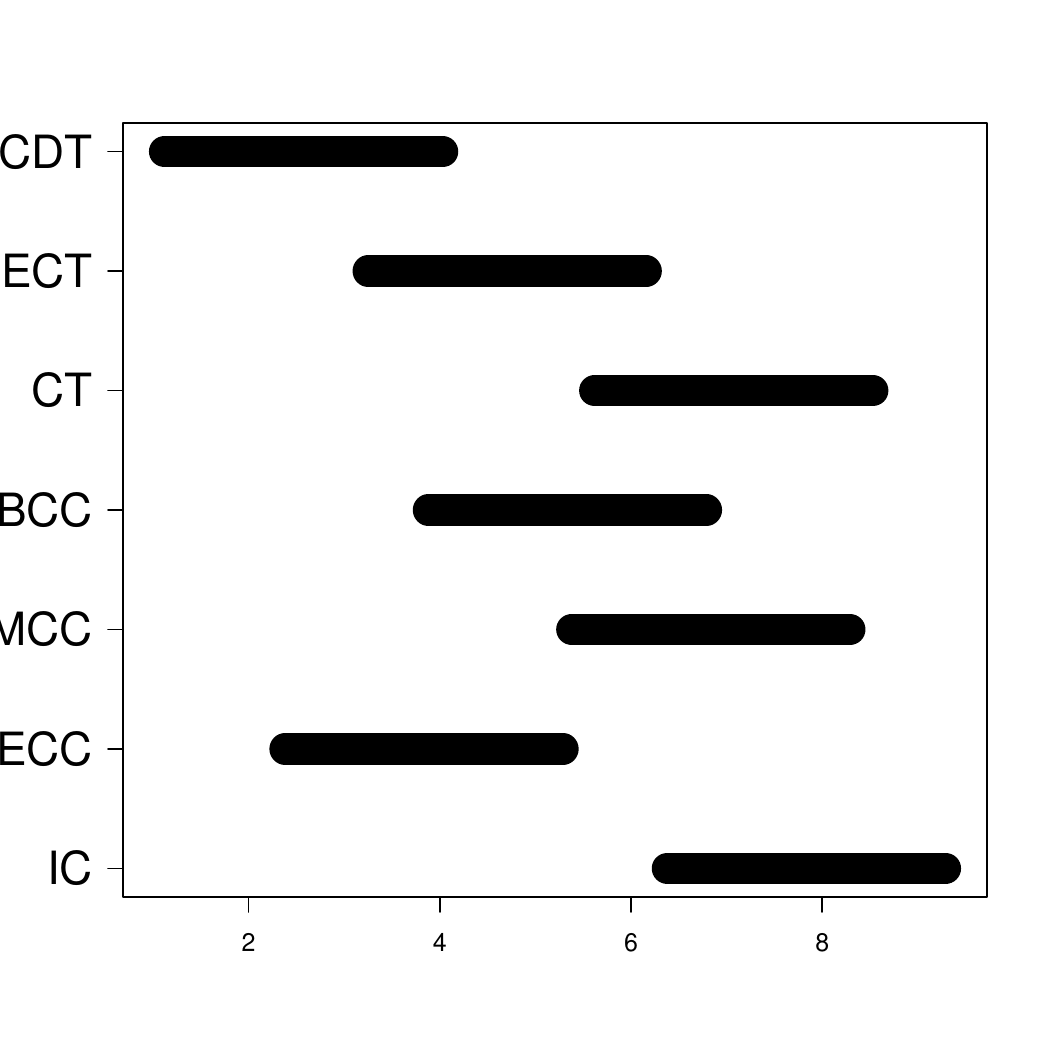}
	}
	\caption{\label{fig:Nemenyi} Results of Nemenyi test, based on \Tab{tab:results} and \Tab{tab:times}, If methods' bars overlap, they can be considered statistically indifferent. The graphs based on time should be interpreted such that higher rank (more to the left) corresponds to slower (i.e., less desirable) times. }
\end{figure}

In order to illustrate the most significant statistical differences between all methods, in \Fig{fig:Nemenyi} we include the results of the Nemenyi test based on \Tab{tab:results} and \Tab{tab:times}. However, note that here we excluded the two rows with DNFs. The Nemenyi test \cite{Nemenyi} rejects the null hypothesis if the average rank difference is greater than the \emph{critical distance} $q_p \sqrt{N_A(N_A+1)/6 N_D}$ over $N_A$ algorithms and $N_D$ datasets, and $q_p$ according to the $q$ table for some $p$ value (we use $p=0.90$). Any method with a rank greater than another method by at least the critical distance, is considered statistically better. In \Fig{fig:Nemenyi}, for each method, we place a bar spanning from the average rank of the method, to this point \emph{plus} the critical distance. Thus, any pair of bars that \emph{do not} overlap correspond to methods that are statistically different in terms of performance. Note that, regarding both training and test running times, \texttt{CT} overlaps considerably with \texttt{IC}, whereas other methods such as \texttt{ECC} and \texttt{MCC} need significantly more training time. We can say that \texttt{ECC} and \texttt{ECT} are statistically stronger than \texttt{IC} and \texttt{CDT}, but not so wrt exact match. \texttt{CT} performs particularly well on the Hamming score, indicating that error propagation is limited, compared to other \texttt{CC} methods.

In the following section we present the framework behind the localization datasets \textsf{Local400} and  \textsf{Local10k} in the tables presented above.

\begin{table*}
	\centering
	\footnotesize
	\caption{\label{tab:times}Time results (seconds). DNF = Did Not Finish (within 24 hours or 2 GB memory).}

    \begin{center}
        Training Time
    \end{center}
    \begin{tabular}{llrrrrrrr}
\hline
Dataset   &  $L$ & \alg{IC}&\alg{ECC}&\alg{MCC}&\alg{EBCC}&\alg{CT}&\alg{ECT}&\alg{CDT} \\
\hline
Music  & $6$               &      1 (3) &      4 (6) &      4 (7) &      2 (5) &      0 (1) &      2 (4) &      1 (2)  \\ 
Scene  & $6$               &      3 (1) &     10 (5) &     28 (7) &      7 (4) &      3 (3) &     10 (6) &      3 (2)  \\ 
Yeast     & $14$             &     11 (3) &     53 (6) &     79 (7) &     45 (5) &      5 (2) &     26 (4) &      5 (1)  \\ 
Medical    & $45$            &      4 (2) &     19 (6) &     67 (7) &     17 (4) &      3 (1) &     19 (5) &      5 (3)  \\ 
Enron     & $53$             &     51 (3) &    207 (6) &    734 (7) &     95 (4) &     24 (1) &    100 (5) &     37 (2)  \\ 
TMC07    & $22$              &  11402 (2) &  48019 (6) &  73433 (7) &  34559 (4) &  10847 (1) &  44986 (5) &  13547 (3)  \\ 
MediaMill   &$101$          &     42 (1) &    347 (6) &   1121 (7) &    238 (5) &     45 (2) &    219 (4) &     55 (3)  \\ 
Delicious  & $983$           &    468 (1) &    DNF     &  18632 (5) &    DNF     &    529 (2) &   2791 (4) &    599 (3)  \\ 
Local400   & $400$     &      2 (1) &     15 (5) &     57 (7) &      8 (3) &      2 (2) &      9 (4) &     31 (6)  \\ 
Local10k   & $10^4$       &      3 (1) &    DNF     &    DNF     &     13 (4) &      3 (2) &     15 (5) &      3 (3)  \\ 
\hline
avg rank &  &  1.80      &  5.75      &  6.78      &  4.22      &  1.70      &  4.60      &  2.80      \\
\hline
\end{tabular}

    \begin{center}
        Test Time
    \end{center}
    \begin{tabular}{llrrrrrrr}
\hline
Dataset   & $L$ & \alg{IC}&\alg{ECC}&\alg{MCC}&\alg{EBCC}&\alg{CT}&\alg{ECT}&\alg{CDT} \\
\hline
Music     &    $6$       &      0 (3) &      1 (7) &      0 (1) &      0 (5) &      0 (4) &      0 (2) &      0 (6)  \\ 
Scene      &   $6$       &      0 (1) &      1 (5) &      0 (2) &      0 (4) &      0 (3) &      2 (6) &      7 (7)  \\ 
Yeast         &  $14$     &      1 (2) &      3 (5) &      3 (6) &      1 (3) &      0 (1) &      2 (4) &      8 (7)  \\ 
Medical       &  $45$     &      4 (2) &     28 (6) &      9 (3) &     26 (5) &      2 (1) &     22 (4) &    141 (7)  \\ 
Enron           & $53$    &      4 (1) &    112 (6) &      8 (3) &     19 (4) &      4 (2) &     45 (5) &    310 (7)  \\ 
TMC07          & $22$     &      7 (2) &     50 (5) &      3 (1) &     42 (4) &      7 (3) &     76 (6) &    534 (7)  \\ 
MediaMill        & $101$   &     15 (1) &    211 (5) &     31 (2) &    143 (4) &     32 (3) &    972 (6) &   5469 (7)  \\ 
Delicious         & $983$  &    167 (1) &    DNF     &    322 (3) &    DNF     &    207 (2) &   7532 (4) &  31985 (5)  \\ 
Local400      &  $400$ &      1 (1) &     17 (6) &      3 (3) &      9 (4) &      1 (2) &     17 (5) &    398 (7)  \\ 
Local10k       &  $10^4$ &      4 (2) &    DNF     &    DNF     &     18 (3) &      4 (1) &     39 (4) &   4228 (5)  \\ 
\hline
avg rank  & &  1.60      &  5.62      &  2.67      &  4.00      &  2.20      &  4.60      &  6.50      \\
\hline
\end{tabular}

\end{table*}

\newpage
\subsection{A Structured Output Prediction Problem}
\label{sec:experiments2}

\begin{figure}[h]
\begin{center}
 \includegraphics[width=0.50\textwidth]{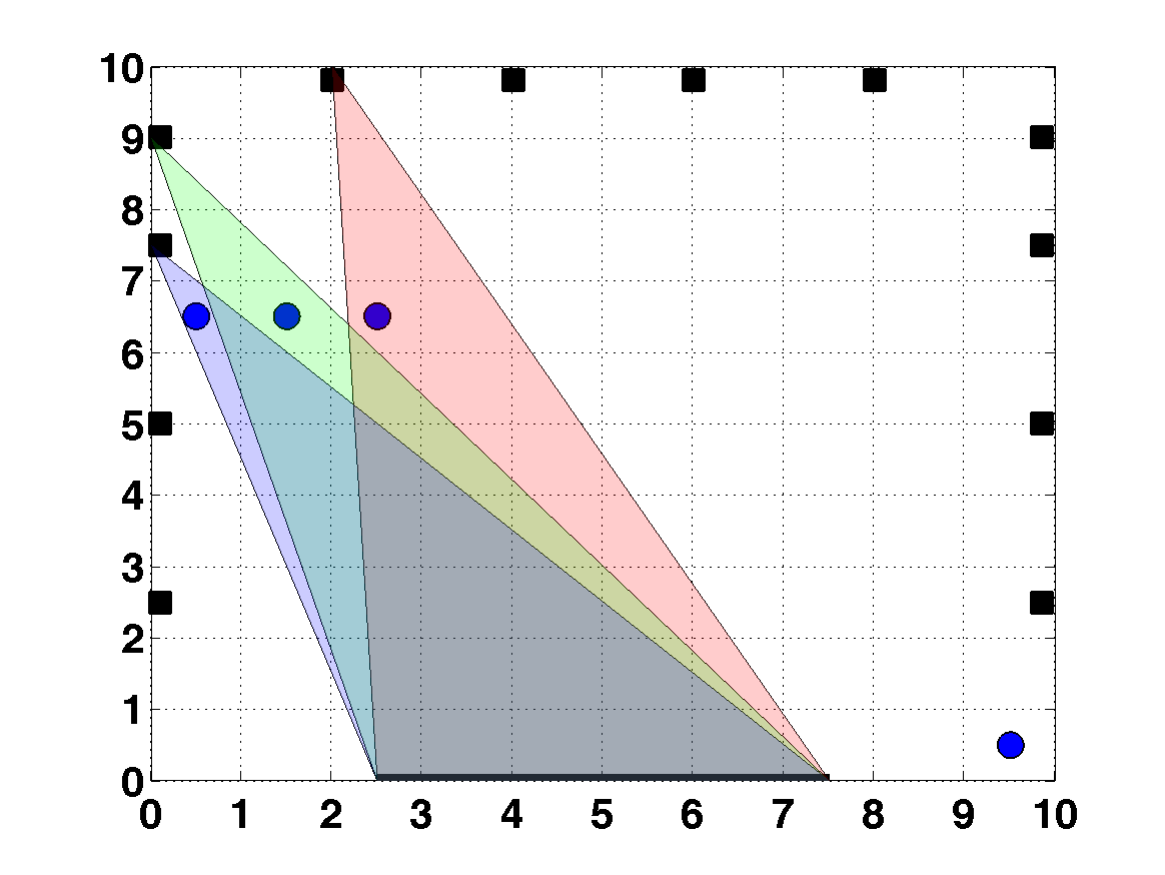} 
\end{center}
\caption{\label{fig:scenL}An $L = 10 \times 10 = 100$ tile localization scenario. $D=12$ light sensors are arranged around the edges of the scenario at coordinates $\s_1,\ldots,\s_D$, and there is a light source between points $\vec{l}_1 = [2.5,0]$ and $\vec{l}_2 = [7.5,0]$ (shown as a thick black line) on the horizontal axis. In this example, three observations are positive ($x_d = 1$). Note that the object in the bottom-right tile ($y_L=1$ in this case) cannot be detected.}
\end{figure}

We investigate the application of \texttt{CT} (and the other MLC methods) to a type of structured output prediction problem:  segmentation for localization.
%
%
In this section we consider a localization application using light sensors, based on the real-world scenario described in \cite{DPF}, where a number of light sensors are arranged around a room for the purpose of detecting the location of a person.
We take a `segmentation' view of this problem, and use synthetic models (which are based on real sensor data) to generate our own observations, thus creating a semi-synthetic dataset, which allows us to easily control the scale and complexity.
\Fig{fig:scenL} shows the scenario.
It is a top-down view of a room with light sensors arranged around the edges, one light source (a window, drawn as a thin rectangle) on the bottom edge and four targets.
Note that targets can only be detected if they come between a light sensor and the light source, so the target in the lower right corner is undetectable.


%

We divide the scenario into $L=W\times W$ square `tiles', representing $Y_1,\ldots,Y_L$. Given an instance $n$,
$$
	\y^{(n)} = \begin{bmatrix}
			y_{1,1}^{(n)} & y_{1,2}^{(n)} & \ldots & \ldots  & y_{1,W}^{(n)} \\
		 \ldots & \ldots & \ldots & \ldots  & \ldots \\
			y_{W,1}^{(n)} & y_{W,2}^{(n)} & \ldots &  \ldots & y_{W,W}^{(n)} \\
		\end{bmatrix},
$$
where $y_{i,j}^{(n)} = 1$ if the $i,j$-th tile (i.e., pixel) is active, and $y_{i,j}^{(n)} = 0$ otherwise, with $i,j \in \{1,\ldots,W\}$.
For the $n$-th instance we have binary sensor observations
$\x^{(n)} = [x_1^{(n)}, \ldots, x_D^{(n)}]$, where $x_d = 1$ if the $d$-th sensor detects an object inside its `detection zone' (shown in colors in \Fig{fig:scenL}).
Otherwise, $x_d=0$.

\subsubsection{Sensor Model}

Consider, for simplicity, a specific instance ${\bf y}=\{y_{i,j}\}_{i,j=1}^{W}$ (in order to avoid here the use of the super index $n$). 
Moreover, let us denote as ${\bf s}_d=[s_{1,d}, s_{2,d}]$ the position of the $d$-th sensor, and $Z_d$ the triangle of vertices ${\bf s}_d$, $\vec{l}_1 = [2.5,0]$ and $\vec{l}_2 = [7.5,0]$ (the corners of the light source).
This triangle $Z_d$ is the ``detection zone'' of the $d$-th sensor.
Now, we define the indicator variable
$$z_{d,i,j}=1 \quad \mbox{ if } \left(i-\frac{1}{2},j-\frac{1}{2}\right) \in Z_d, \quad  z_{d,i,j}=0 \quad \mbox{ if }   
	\left(i-\frac{1}{2},j-\frac{1}{2}\right)  \notin Z_d,$$
where $ \left(i-\frac{1}{2},j-\frac{1}{2}\right)$ is the middle point of the $(i,j)$-th pixel (tile), whose vertices are $(i,j)$, $(i-1,j-1)$, $(i-1,j)$ and $(i,j-1)$ (for $i,j=2,\ldots, W$).
Next, we define the variable
\begin{equation}
	c_d = \sum_{i=1}^W\sum_{j=1}^W y_{i,j}z_{d,i,j},
\label{eq:cd}
\end{equation}
which corresponds to the number of active tiles/pixels inside the triangle $Z_d$ associated to the $d$-th sensor.
The likelihood function for the $d$-th sensor is then given by
\begin{equation}
	p(x_d=1|\y)=p(x_d=1|c_d)= \left\{ 
		\begin{array}{l l}
		    \epsilon_2,    & \quad c_d=0;  \\
		    1- \epsilon_1,     & \quad c_d=1;  \\
            1 - \epsilon_1 \exp[-0.1 (c_d-1)], & \quad c_d > 1; \\
		\end{array} 
	\right.
	\label{eq:observation_model}
\end{equation}
and $p(x_d=0|\y) = 1 - p(x_d=1|\y)$; where $\epsilon_1 = p(x_d=0| c_d=1) = 0.15 < 0.5$ is the false \emph{negative} rate and $\epsilon_2 = p(x_d=1| c_d=0 ) = 0.01<0.5$  is the false \emph{positive} rate. 



\subsubsection{Generation of Artificial Data}

\Fig{fig:scenL} shows a low dimensional scenario ($L=100, W=10$), for the purpose of a clear illustration, but we consider datasets with much higher levels of segmentation (namely \data{Local400}, where $L=400$, and \data{Local10k}, where $L=10,000$ -- see \Tab{tab:datasets}) to compare the performance of several MOC techniques on this problem.
Given a scenario with $L = W \times W$ tiles, $D$ sensors and $N$ observations, we generate the synthetic data, $(\x^{(n)},\y^{(n)})_{n=1}^{N}$, as follows:
\begin{enumerate}
	\item Start with an `empty' $\y$, i.e., $y_{i,j}=0$ for $i,j = 1, \ldots, W$.
	\item \label{step:2} Set $y_{i,j}=1$ for relevant tiles to create a rectangle of width $W/8$ and height $2$ starting from some random point $y_{i,j}$.
	\item \label{step:3} Create a $2 \times 2$ square in the corner \emph{furthest from} the rectangle.
	\item Generate the observations according to \Eq{eq:observation_model}.
	\item Add dynamic noise in $\y$ by flipping $L/100$ pixels uniformly at random.
\end{enumerate}

Any MLC method can be applied to this problem, to infer the binary vector $\y^*$, which encodes the presence of blocking-light elements in the room, given the vector $\x^*$ of measurements from the light sensors.
Finally, we also consider that each sensor provides $M$ observations $\{x_{d,k}\}_{k=1}^M\subset\{0,1\}^M$ given the same $\y$. 
\begin{code}
    \caption{\label{code:BasicInference}MAP inference using the sensor model}
\removelatexerror
\begin{algorithm}[H]
\footnotesize
    \SetKwInOut{Input}{input}
    \SetKwInOut{Output}{output}
    \Input{
       $\{x_{d,k}\}_{k=1,d=1}^{M,D}$ (measurements), $\{{\bf s}_d\}_{d=1}^D$ (sensor positions), $\vec{l}_1$ and $\vec{l}_2$ (light source location).
    }
    \Begin{
         \begin{enumerate}
   \item    Initialize ${\hat y}_{i,j}=0.5$ for $i,j=1,\ldots, W$.  
   \item    \For{$d=1,\ldots,D$;}{
      \begin{enumerate}
      	\item Calculate the detection triangle $Z_d$.
      	\item If  $\hat{\theta}_d=\frac{1}{M}\sum_{k=1}^M x_{d,k}\leq 0.5$, then set ${\hat c}_d=0$.
      	\item Otherwise, if $\hat{\theta}_d=\frac{1}{M}\sum_{k=1}^M x_{d,k}> 0.5$, then set ${\hat c}_d=1$.
      	\item If ${\hat c}_d=0$, then set ${\hat y}_{i,j}=0$ for all $i,j \in Z_d$.   
      \end{enumerate}
        }
\item For all $d$ such that ${\hat c}_d=1$ and for all $i,j\in Z_d$, check if the decision is still ${\hat y}_{i,j}=0.5$. Then, set ${\hat y}_{i,j}=1$.
 \item The remaining tiles with ${\hat y}_{i,j}=0.5$ correspond to ``shadow'' zones, where we leave ${\hat y}_{i,j}=0.5$.
\end{enumerate}    
    }
    \Output{ ${\hat y}_{i,j}$ for $i,j=1,\ldots, W$. 
        }
\end{algorithm}
\end{code}

\subsubsection{Maximum A Posteriori (MAP) Estimator}

Given the likelihood function of \Eq{eq:observation_model}, and considering a uniform prior over each variable $c_d$, the posterior w.r.t.\ the $d$-th triangle is 
$$
p(c_d|x_d) \propto  p(x_d|c_d)=p(x_d|\y).
$$
If we also assume independency in the received measurements, the posterior density $p(c_1,\ldots,c_D|\x)$ can be expressed as follows
\begin{equation}
p(c_1,\ldots,c_D|\x){\color{black} = \prod_{d=1}^{D} p(c_d|\x)}\propto \prod_{d=1}^{D} p(x_d|c_d)= \prod_{d=1}^{D} p(x_d|\y).  
\end{equation}
We are interested in studying $p(y_{i,j}|{\bf x})$ for $1 \le i,j \le W$, but we can only compute the posterior distribution of the variables $\{c_1,\ldots c_D\}$, which depend on $y_{i,j}$ through \Eq{eq:cd}.
Making inference directly on $y_{i,j}$ using the posterior distribution $p(c_1,\ldots,c_d|\x)$ is not straightforward.
Let us address the problem in two steps.
First, the measurements received by each sensor, $x_d$, can be considered as Bernoulli trials: if $c_d=0$, then $x_d=1$ with probability $\theta_d=\epsilon_2$; if $c_d\geq 1$, then $x_d=1$ with success probability $\theta_d=1-\epsilon_1$.
%
Now, given $M$ measurements for each sensor $\{x_{d,k}\}_{k=1}^M\subset\{0,1\}^M$ and uniform prior density over $\theta_d$, the MAP estimator of $\theta_d$ is given by
$$
{\hat{\theta}}_d=\frac{1}{M}\sum_{k=1}^M x_{d,k}.
$$
Then, if ${\hat{\theta}}_d\leq 0.5$ we decide $\hat{c}_d=0$.
Otherwise, if $\hat{\theta}_d> 0.5$, we estimate $\hat{c}_d\geq 1$.
Considering a uniform prior over the pixels $y_{i,j}$, a simple procedure to estimate $\y$ from $\{\hat{c}_1,\ldots \hat{c}_D\}$  is the one described in Algorithm \ref{code:BasicInference}.

\subsubsection{Classifier Trellis vs MAP Estimator}

Results for \texttt{CT} are already given in \Tab{tab:results} (predictive performance) and \Tab{tab:times} (running time).
Results in \Tab{tab:results} illustrate the robustness of the \texttt{CT} algorithm to address multi-output classification in several scenarios.
Beyond the training set, no further knowledge about the underlying model is needed to achieve remarkable classification performance.
To emphasize this property of \texttt{CT}, we now compare it to the MAP estimator presented above, which exploits a perfect knowledge of the sensor model.

Table \ref{tab:ResultsLuca} shows the results using Algorithm \ref{code:BasicInference} with $D=30$ sensors and different values of $W$ (i.e., the grid precision).
The corresponding results obtained by CT are provided in Table \ref{tab:ResultsJesse}.
A detailed discussion of these results is provided at the end of the next Section.
However, let us remark that increasing the number of tiles (i.e., $W$) for a given number of sensors $D$ makes the problem harder, as a finer resolution is sought.
This explains the decrease in performance seen in the tables as $W$ increases.

\begin{table*}
    \centering
    \caption{\label{tab:ResultsLuca} Results using Algorithm \ref{code:BasicInference} with $D=30$ sensors. }
  \footnotesize
	\begin{tabular}{rcc}
        \hline
		{\bf Measure} & $W=20$ & $W=100$ \\ 
				\hline
               Accuracy & 0.523  & 0.141 \\
               Hamming score &0.857  & 0.795\\
				Times (total, s) & 1      & 3     \\
          \hline		
           \end{tabular}
\end{table*}
 
 \begin{table*}
    \centering
    \caption{\label{tab:ResultsJesse} Results using CT ($D=30$ sensors). }
  \footnotesize
	\begin{tabular}{rcc}
        \hline
        {\bf Measure} & $W=20$ & $W=100$ \\ 
		\hline		
               Accuracy & 0.542 & 0.133 \\
               Hamming score & 0.969 & 0.968 \\	
				Time (total, s) & 3      & 7  \\
          \hline		
           \end{tabular}
\end{table*}

\section{Discussion}
\label{sec:conclusions}

As in most of the multi-label literature, we found that independent classifiers consistently under-perform, thus justifying the development of more complex methods to model label dependence.
However, in contrary to what much of the multi-label literature suggests, greater investments in modelling label dependence do not always correspond to greater returns.
In fact, it appears that many methods from the literature have been over-engineered.
Our small experiment in \Tab{tab:2} suggests that none of the approaches we investigated were particularly dominant in their ability to uncover structure with respect to predictive performance.
Indeed, our results indicate that none of the techniques is significantly better than another. Using \texttt{ECC} is a `safe bet' in terms of high accuracy, since it models long term dependencies with a fully cascaded chain; also noted previously (e.g,. \cite{ECC2,PCC}).
%
%
%
%
In terms of \texttt{EBCC} (for which we elected to represent methods that uncover a structure), there was no clear advantage over the other methods, and surprisingly also no clear difference between searching for a structure based on marginal dependence versus conditional label dependence.
%
%
This makes it more difficult to justify computationally complex expenditures for modelling dependence on the basis of improved accuracy; particularly so for large datasets, where the scalability is crucial.

We presented a classifier trellis (CT) as an alternative to methods that model a full chain (as \texttt{MCC} or \texttt{ECC}) or methods that unravel the label graphical model structure  from scratch, such as \texttt{BCC}. Our approach is systematic, we consider a fixed structure in which we place the labels in an ordered procedure according to easily computable mutual information measures, see \Code{code:CT}. An ensemble version of \texttt{CT} performs particularly well on exact match but,  surprisingly, it does not perform much stronger than \texttt{CT} as we expected in the beginning. It does not perform as strong overall as \texttt{ECC} (although there is no statistically significant difference), but is much more scalable, as indicated in \Tab{tab:complexity}. 

The \texttt{CT} algorithm then emerges as a powerful MLC algorithm, able to excellent performance (specially in terms of average number of successfully classified labels) with near $\texttt{IC}$ running times.  Through the Nemenyi test, we have shown the statistical similitude between the classification outputs of \texttt{(E)CT}  and \texttt{MCC}/\texttt{ECC}, proving that our approach based on the classifier trellis captures the necessary inter-label dependencies to achieve high performance classification. Moreover, we have not analyzed yet the impact that the trellis structure chosen has in the \texttt{CT} performance. In future work, we intend to experiment with trellis structures with different degrees of connectedness.

\section{Acknowledgements}

This work was supported by the Aalto University AEF research programme; by the Spanish government's (projects 
projects 'COMONSENS', id. CSD2008-00010, 'ALCIT', id. TEC2012-38800-C03-01, 'DISSECT', id. TEC2012-38058-C03-01); by Comunidad de Madrid in Spain (project 'CASI-CAM-CM', id. S2013/ICE-2845); and by and by the ERC grant 239784 and AoF grant 251170.

\bibliographystyle{plain}
\bibliography{MCMC}

\appendix

\section{Graphical Model Structure Learning: \textsc{FS} vs. \textsc{LEAD}}\label{sec:appendixA}

In order to delve deeper into the issue of structure learning, we generated a synthetic dataset, where the underlying structure is known.
The synthetic generative model is as follows.
For the feature vector, we consider a $D$-dimensional independent Gaussian vector ${\bf x}\in\mathbb{R}^D$, where $x_d \sim \mathcal{N}(0,1)$ for $d = 1, \ldots, D$.
Let $\mathbf{w}_{\ell}$ ($\ell = 1, \ldots, L$) be a binary $D$-dimensional vector containing exactly $T$ ones (and thus $D-T$ zeros), and let us assume that we have a directed acyclic graph between the labels in which each label has at most one parent.
Both the  vectors $\mathbf{w}_{\ell}$ and the dependency label graph are generated uniformly at random.
Given the value of its parent label, $y_{\pa{\ell}}\in\{-1,1\}$, the following probabilistic model is used to generate the $\ell$-th label $y_{\ell}$:
\begin{equation}
\label{toymodel}
	y_{\ell} = \begin{cases}
		+1, & T^{-1/2} \mathbf{w}_{\ell}^{\top} {\bf x} + \epsilon_{\ell} \geq \delta;\\
		-1, & \text{otherwise};
	\end{cases}
\end{equation}
where $\delta$ is a real constant and $\epsilon_{\ell} \sim \mathcal{N}(\alpha ~y_{\pa{\ell}}, \sigma^2)$, with $\alpha\in\mathbb{R}$ and $\sigma\in\mathbb{R}^{+}$.
Note that, according to the model, $y_{\ell}|y_{\pa{\ell}}$ is a Bernoulli random variable that takes value $1$ with average probability
\begin{align*}
P(y_{\ell}=+1|y_{\pa{\ell}}=+1) = Q\left(\frac{\delta-\alpha~ y_{\pa{\ell}}}{\sqrt{1+\sigma^2}}\right),
\end{align*}
where $Q(x)=1-\Phi(x)$ and $\Phi(x)$ is the cumulative distribution function of the normal Gaussian distribution.
%
Consequently, with $\alpha$ and $\sigma^2$ we control the likelihood of $y_{\ell}$ being equal to its parent $y_{\pa{\ell}}$, thus modulating the complexity of inferring such dependencies by using the \textsc{fs} and \textsc{lead} methods.

\begin{figure}
	\centering
	\subfloat[test][Ground Truth]{
		\includegraphics[scale=0.3]{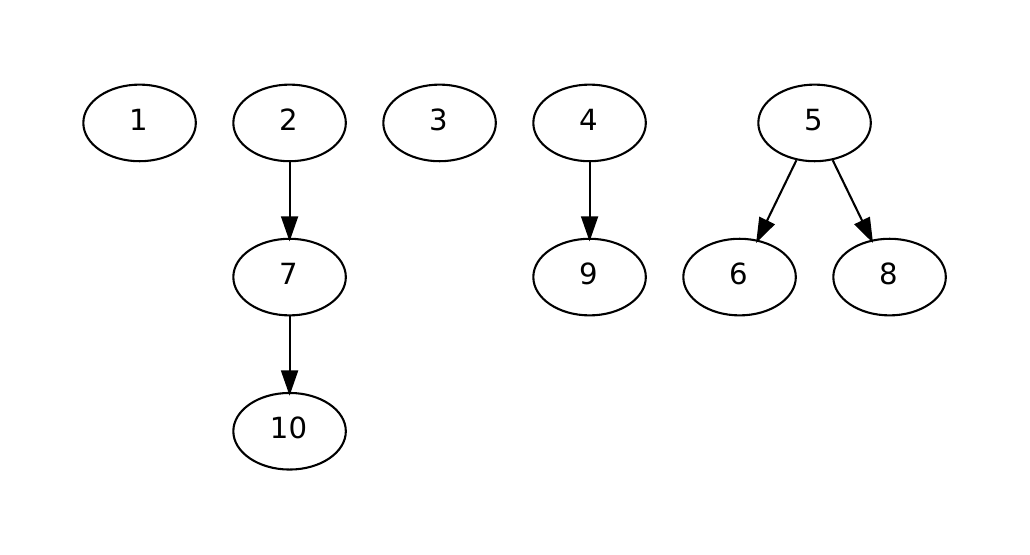}
	}
	\subfloat[test][\textsc{fs}]{
		\includegraphics[scale=0.3]{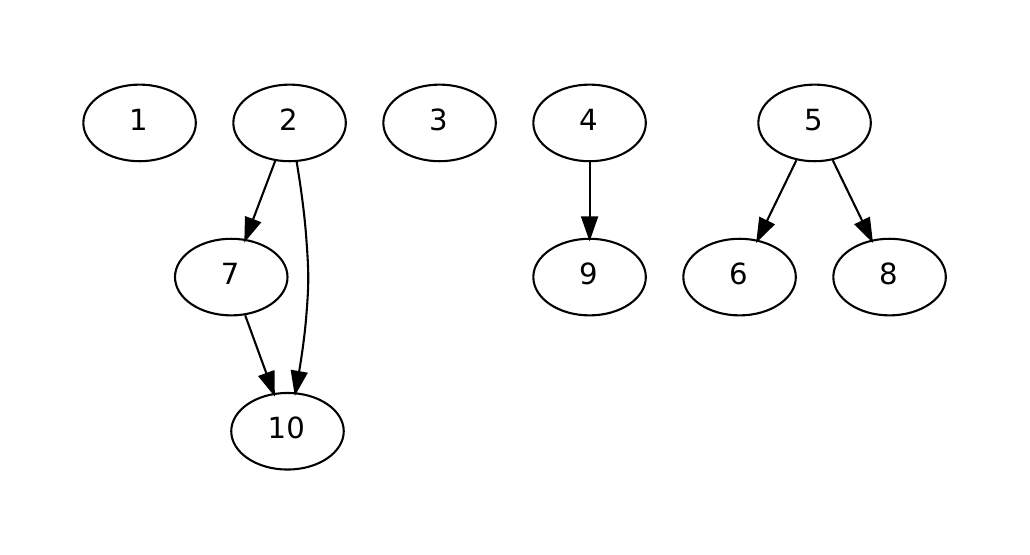}
	}
	\subfloat[test][\textsc{lead}]{
		\includegraphics[scale=0.3]{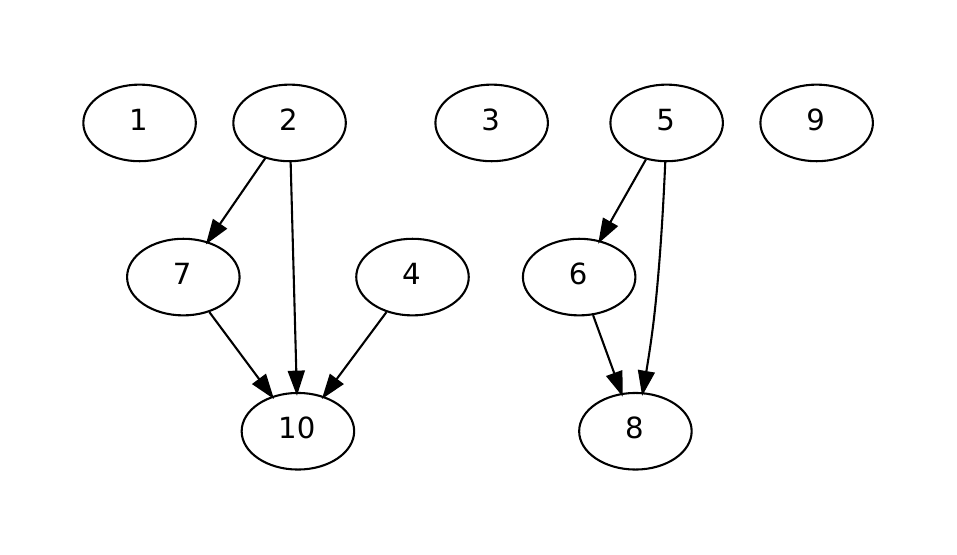}
	}
	\qquad
	\subfloat[test][Ground Truth]{
		\includegraphics[scale=0.3]{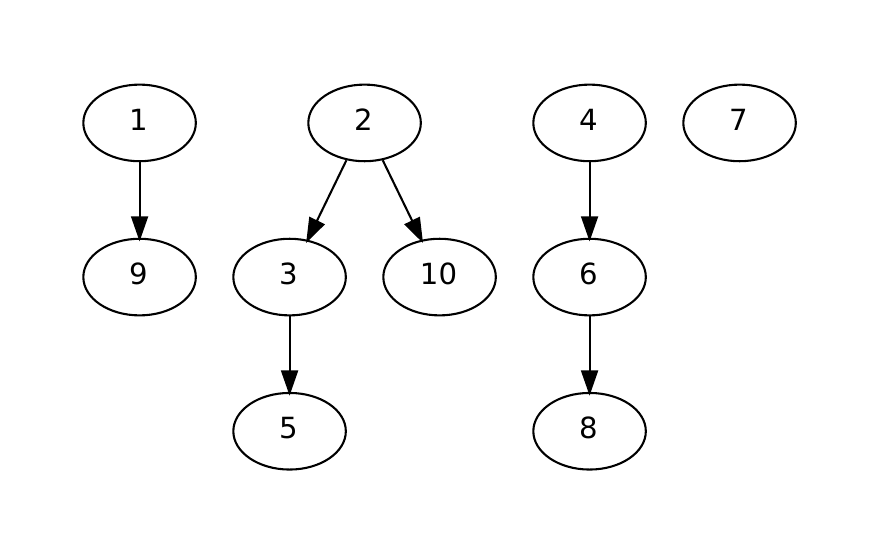}
	}
	\subfloat[test][\textsc{fs}]{
		\includegraphics[scale=0.3]{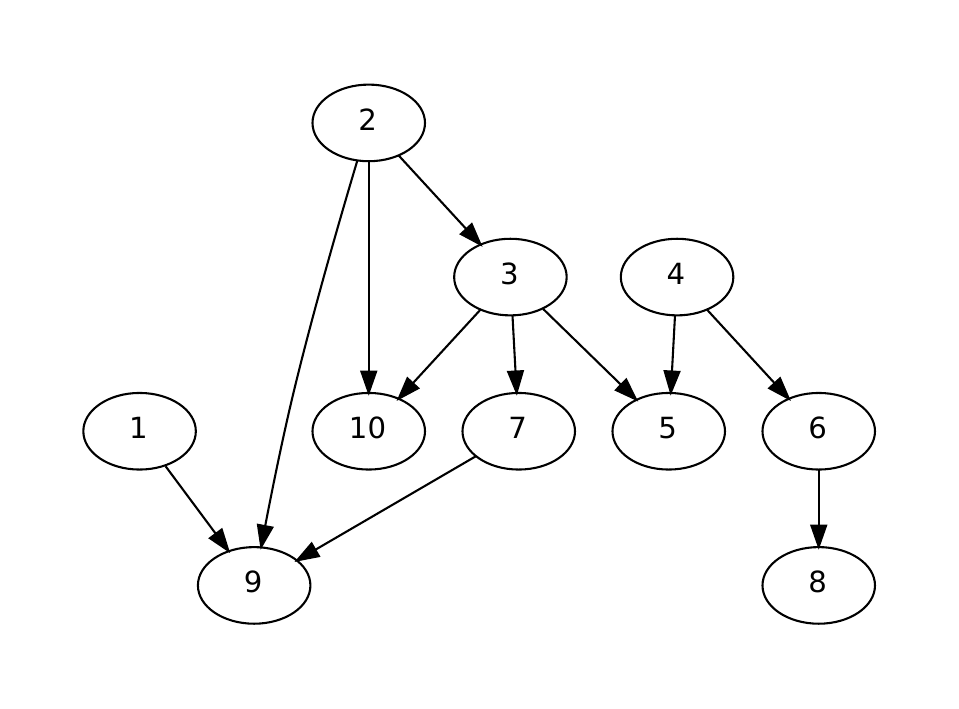}
	}
	\subfloat[test][\textsc{lead}]{
		\includegraphics[scale=0.3]{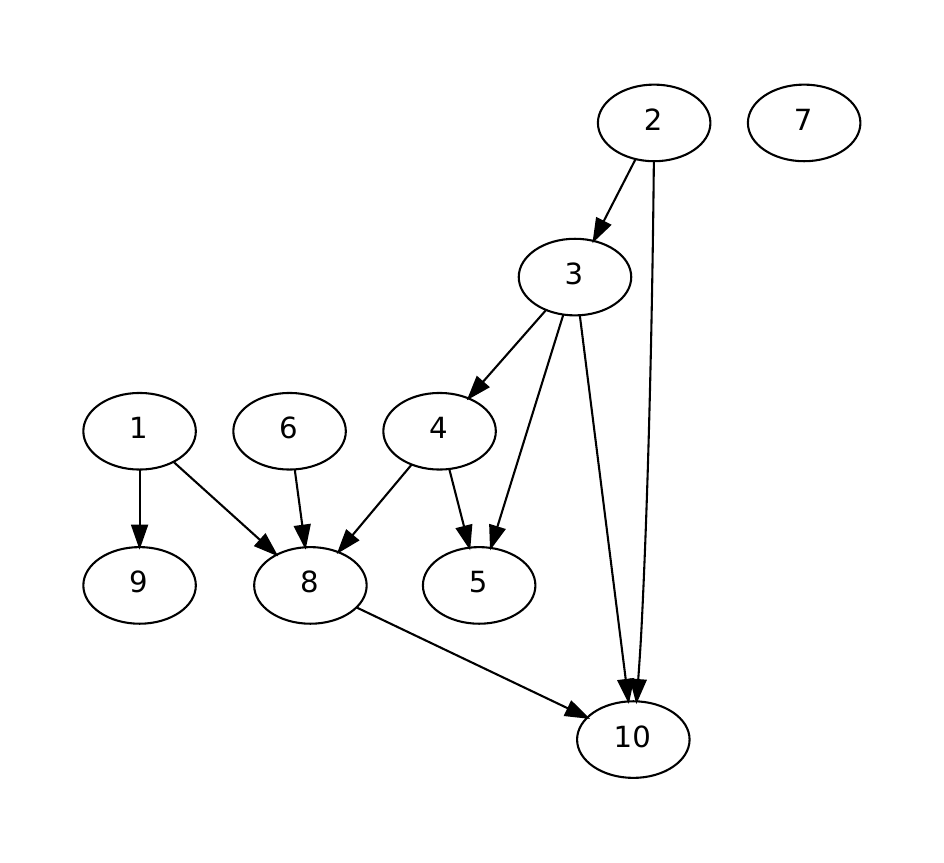}
	}
	\qquad
	\subfloat[test][Ground Truth]{
		\includegraphics[scale=0.3]{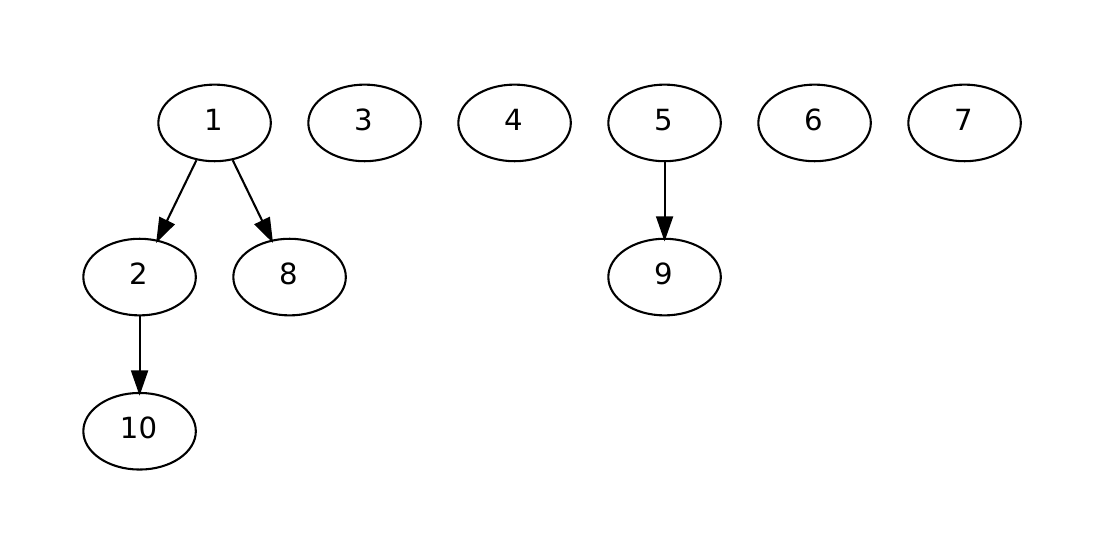}
	}
	\subfloat[test][\textsc{fs}]{
		\includegraphics[scale=0.3]{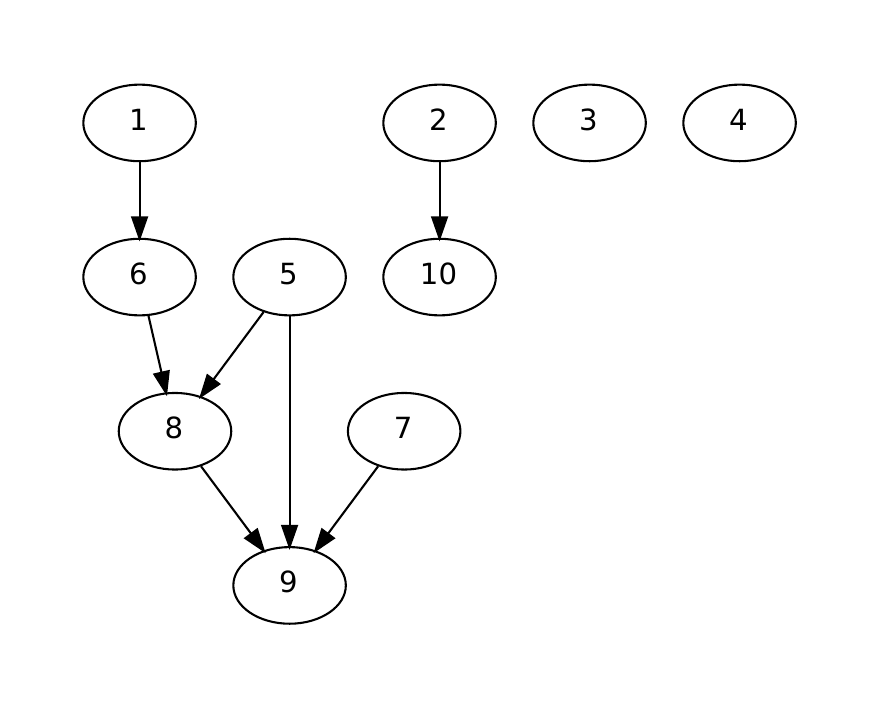}
	}
	\subfloat[test][\textsc{lead}]{
		\includegraphics[scale=0.3]{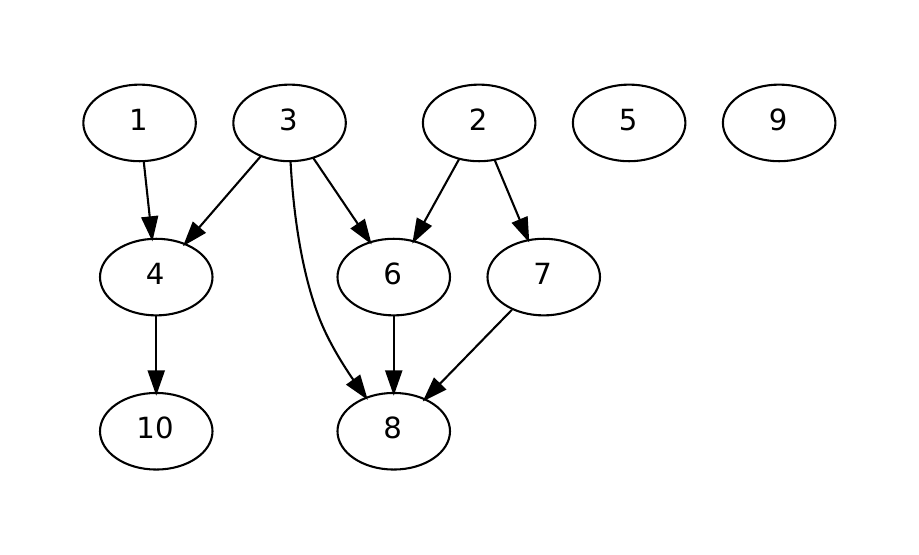}
	}
	\begin{center}
	\begin{tabular}{llll}
		\hline
		   parameter & easy & medium & hard \\
		\hline
		$\alpha$ & 1 & 0.5 & 0.25 \\
		$\sigma^2$  & 1 & 2   &    5 \\
		\hline
	\end{tabular}
	\end{center}
	\caption{\label{fig:toyrecon}Ground-truth graphs of the synthetic dataset (left) -- easy, medium, and hard according to the table -- and their reconstruction found by the \textsc{fs} strategy 
		\cite{BNFS} (middle) and \textsc{lead} strategy \cite{LEAD} (right).}
\end{figure}

In \Fig{fig:toyrecon} we show three examples of synthetically generated datasets, in terms of their ground truth structure and the structure discovered using the \textsc{fs} and \textsc{lead} methods, for three different scenarios: `easy' ($\alpha=1$, $\sigma^2=1$), `medium' ($\alpha=0.5$, $\sigma^2=2$), and `hard' ($\alpha=0.25$, $\sigma^2=5$) datasets.
Recall that we use a \key{mutual information} matrix for both methods, with the difference being that the \textsc{lead} matrix is based on the \key{error frequencies} rather than the label frequencies. 
Visually it appears that both \textsc{fs} and \textsc{lead} are able to discover the original structure, relative to the difficulty of the dataset.
There appears to be a small improvement of \textsc{fs} over \textsc{lead}.
This is confirmed in a batch analysis using the F-measure of 10 random datasets of random difficulty ranging between `easy' and `hard': \textsc{fs} gets $0.278$ and \textsc{lead} gets $0.263$. 
A more in depth comparison, taking into account varying numbers of labels and features, is left for future work.

%
%
\section{Approximate inference via Monte Carlo}
\label{sec:inference}

A better understanding of the MLC/MOC approaches described in Section \ref{sec:prior} and the novel scheme introduced in this work can be achieved by describing the Monte Carlo (MC) procedures used to perform approximate inference over the graphical models constructed to approximate $p(\y|\x)$.

%
Given a probabilistic model for the conditional distribution $p(\y|\x)$ and a new test input $\xtest$, the goal of an MC scheme is generating samples from $p(\y|\xtest)$ that can be used to estimate its mode (which is the MAP estimator of $\y$ given $\xtest$), the marginal distribution per label (i.e., $p(y_{\ell}|\xtest)$ for $\ell=1\ldots,L$) or any other relevant statistical function of the data.

\subsection{Bayesian networks}
\label{sec:BN}

In a directed acyclic graphical model, the probabilistic dependencies between variables are ordered.
For instance, in the \texttt{CC} scheme $p(\y|\x)$ factorizes according to \Eq{eq:chain}.
If it is possible to draw samples directly from each conditional density $p(y_{\ell}|y_{1:\ell-1},{\bf x})$, then exact sampling can be performed in a simple manner.
For $i=1,\ldots,N_s$ (where $N_s$ is the desired number of samples), repeat
\begin{eqnarray*}
	\nonumber
	y_1^{(i)} & \sim & p(y_1|{\bf x})),\\
	y_2^{(i)} & \sim & p(y_2|y_1^{(i)},{\bf x})),\\
	 & \vdots & \\
	y_L^{(i)} & \sim & p(y_L|y_{1:L-1}^{(i)},{\bf x}).
\end{eqnarray*}
For Bayesian networks that are not fully-connected, as in \texttt{BCC}, the procedure is similar.
Each sampled vector, $\y^{(i)} = [y_1^{(i)}, y_2^{(i)}, \ldots, y_L^{(i)}]$ for $i=1,\ldots,N_s$, is obtained by drawing each individual component independently as $y_{\ell}^{(i)} \sim p(y_{\ell}|\y^{(i)}_{\pa{\ell}})$ for $\ell=1,\ldots,L$.

\subsection{Markov networks}
\label{sec:MRF}

In an undirected graphical model (like that of a \texttt{CDN}), 
%
exact sampling is generally unfeasible.
However, a Markov Chain Monte Carlo (MCMC) technique that is able to generate samples from the target density $p({\bf y}|{\bf x})$ can be implemented.
%
%
Within this class, \key{Gibbs Sampling} is often the most adequate approach.
Let us assume that the conditional distribution $p(\y|\x)$ factorizes according to an undirected graphical model as in \Eq{eq:undirected}.
Then, from an initial configuration ${\bf y}^{(0)}=[y_1^{(0)},\ldots,y_L^{(0)}]$, repeat for $t=1, \ldots, T$:
\begin{eqnarray*}
	y_1^{(t)} & \sim & p(y_1|y_{1}^{(t-1)},y_{2}^{(t-1)},...,y_{L}^{(t-1)}, {\bf x}), \\
	y_2^{(t)} & \sim & p(y_2|y_1^{(t)},y_{2}^{(t-1)},...,y_{L}^{(t-1)}, {\bf x}), \\
	& \vdots & \\
	y_L^{(t)} & \sim &  p(y_L|y_1^{(t)},y_{2}^{(t)},...,y_{L}^{(t-1)},{\bf x}),
\end{eqnarray*}
where each label can be sampled by conditioning just on the neighbors in the graph, as seen from \Eq{eq:Markov}.
Thus,
\begin{align*}
	y_{\ell}^{(t)} \sim p(y_{\ell}|\{y_u^{(t)}: y_u \in \y_{\ne{\ell}}, u < \ell\},
		\{y_m^{(t-1)}: y_m \in \y_{\ne{\ell}}, m > \ell\},{\bf x}),
\end{align*}
which can be simply denoted as $y_{\ell}^{(t)}\sim p(y_{\ell}|\y_{\ne{\ell}}^{(t)})$, with $\y_{\ne{\ell}}^{(t)}$ denoting the state of the neighbors of $y_{\ell}$ at time $t$.



Following this approach, the state of the chain $\y^{(t)} = [y_1^{(t)},\ldots, y_L^{(t)}]$ can be considered a sample from $p(\y|\x)$
after a certain ``burn-in'' period, i.e., for $t>T_c$.
Thus, samples for $t<T_c$ are discarded, whereas samples for $t>T_c$ are used to perform the desired inference task.
Two problems associated to MCMC schemes are the difficulty in determining exactly when the chain has converged and the correlation among the generated samples (unlike the schemes in \Sec{sec:BN}, which produce i.i.d.\ samples).
%


%
%



\end{document}